\documentclass{article}

\usepackage[preprint]{neurips_2024}

\usepackage[utf8]{inputenc} 
\usepackage[T1]{fontenc}    
\usepackage{hyperref}       
\usepackage{url}            
\usepackage{booktabs}       
\usepackage{amsfonts}       
\usepackage{nicefrac}       
\usepackage{microtype}      
\usepackage{xcolor}         
\usepackage{graphicx}       
\usepackage{amssymb}        
\usepackage{tcolorbox}      

\setcitestyle{numbers,square,comma}
\title{Large Language Models are Biased Reinforcement Learners}

\author{%
  William M. Hayes \\
  Binghamton University\\
  \texttt{whayes2@binghamton.edu} \\
  \And
  Nicolas Yax \\
  École Normale Supérieure \\
  \texttt{nicolas.yax@ens.psl.eu} \\
  \And
  Stefano Palminteri \\
  École Normale Supérieure \\
  \texttt{stefano.palminteri@ens.fr} \\
}

\begin{document}

\maketitle

\begin{abstract}
  In-context learning enables large language models (LLMs) to perform a variety of tasks, including learning to make reward-maximizing choices in simple bandit tasks. Given their potential use as (autonomous) decision-making agents, it is important to understand how these models perform such reinforcement learning (RL) tasks and the extent to which they are susceptible to biases. Motivated by the fact that, in humans, it has been widely documented that the value of an outcome depends on how it compares to other local outcomes, the present study focuses on whether similar value encoding biases apply to how LLMs encode rewarding outcomes. Results from experiments with multiple bandit tasks and models show that LLMs exhibit behavioral signatures of a relative value bias. Adding explicit outcome comparisons to the prompt produces opposing effects on performance, enhancing maximization in trained choice sets but impairing generalization to new choice sets. Computational cognitive modeling reveals that LLM behavior is well-described by a simple RL algorithm that incorporates relative values at the outcome encoding stage. Lastly, we present preliminary evidence that the observed biases are not limited to fine-tuned LLMs, and that relative value processing is detectable in the final hidden layer activations of a raw, pretrained model. These findings have important implications for the use of LLMs in decision-making applications.
\end{abstract}

\section{Introduction}

In recent years, large language models (LLMs) have captured the attention of the media, industry, and the scientific community \citep{Fowler_2024, Newport_2023, Ornes_2024}. LLMs are essentially massive neural networks, based on the transformer architecture \citep{vaswani2017attention}, that are trained on a vast amount of text data to predict the next token in a sequence. Yet, LLMs have been shown to have emergent abilities that extend beyond next-token prediction \citep{wei2022emergent}. For example, LLMs can translate from one language to another, answer questions, unscramble words, classify text, solve math problems, and perform many other language-based tasks at a high level.

These tasks are accomplished through in-context learning \citep{brown2020language}: using only the contextual information contained in the text input to accomplish a novel task, without any gradient updates or fine-tuning. In-context learning gives LLMs the human-like ability to perform new tasks from just a few examples (few-shot learning) or from instructions alone (zero-shot learning) \citep{kojima2022large}. Larger LLMs with hundreds of billions or even trillions of parameters are particularly good at in-context learning, and performance tends to increase with the number of examples provided in the prompt \citep{brown2020language, wei2022chain}.

As their in-context learning abilities continue to improve, LLMs are being increasingly applied to solve real-world problems \citep{liang2023code, sha2023languagempc, thirunavukarasu2023large}. Of particular interest, LLMs are now being used to augment reinforcement learning (RL) systems \citep{cao2024survey}. RL encompasses sequential decision problems in which agents interact with the environment through trial-and-error to maximize reward \citep{sutton2018reinforcement}. Within the RL framework, LLMs can play many different roles. They may assist in processing information from the environment or make decisions based on the observations and instructions provided \citep{cao2024survey}. Given the applications of LLMs in RL and their potential to function as autonomous agents \citep{park2023generative,wang2024survey,xi2023rise}, it is important to understand how LLMs use in-context learning to make reward-maximizing choices in sequential decision problems.

To investigate the underlying behavioral mechanisms of LLMs, researchers have begun treating the models as participants in psychological experiments \citep{hagendorff2023machine}. This “machine psychology” approach has already proven successful at revealing similarities and differences between human and LLM decision making in multiple domains \citep{aher2023using,binz2023using,chen2023emergence,coda2023inducing,coda2024cogbench,horton2023large,schubert2024context,yax2023studying}. Traditional methods for evaluating LLMs rely on overall performance benchmarks across a multitude of tasks \citep{hendrycks2020measuring, srivastava2022beyond, zheng2024judging}, focusing solely on how well the models perform in each task. In contrast, machine psychology focuses on \textit{how} LLMs accomplish the task, using a combination of carefully designed experiments and computational cognitive modeling to characterize their behavior patterns \citep{coda2024cogbench, schubert2024context}.

\begin{figure}
  \centering
  \includegraphics[width=5.5in]{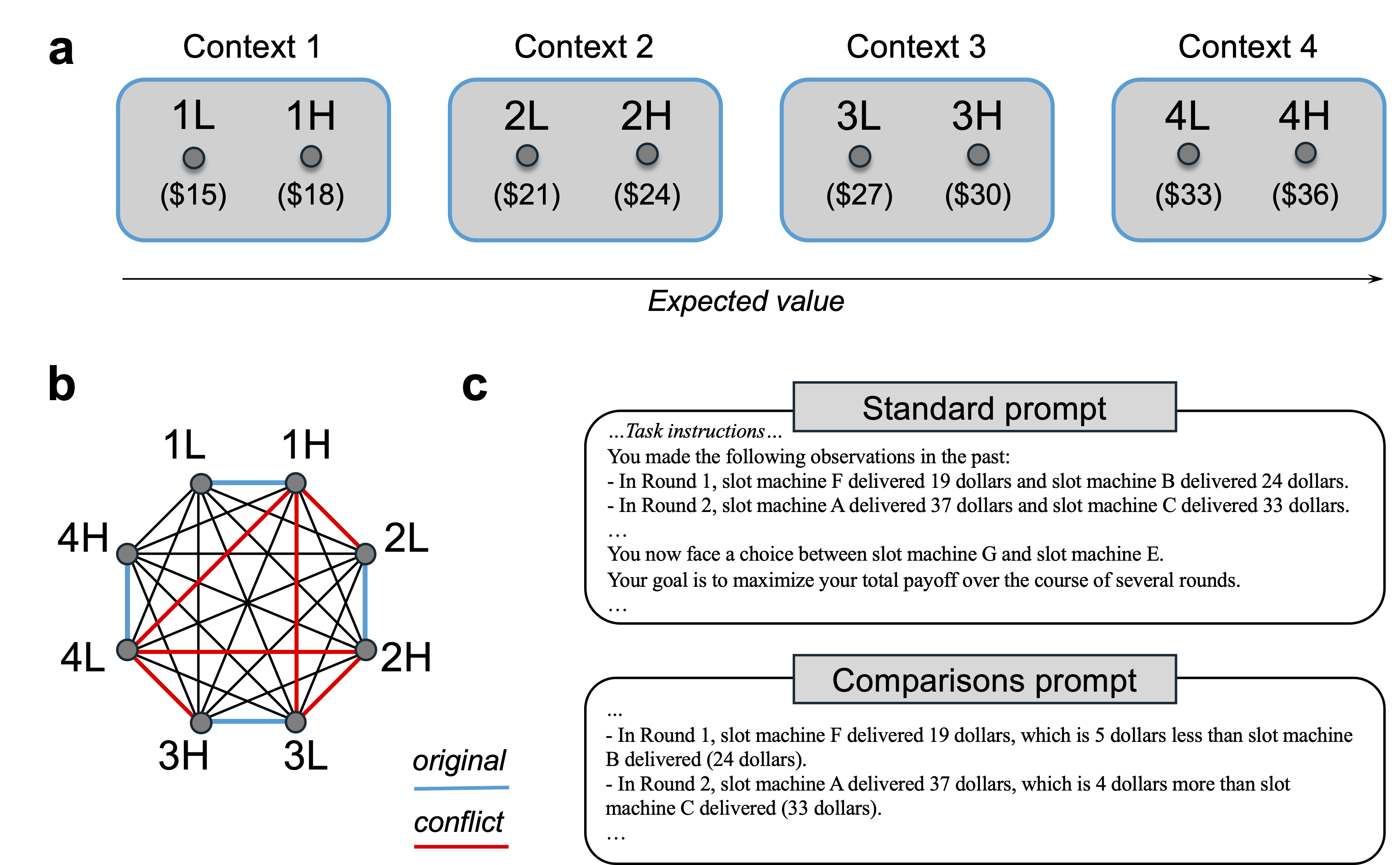}
  \label{Fig1}
  \caption{(a) A bandit task with eight options grouped into four contexts. Each context has a lower value option and a higher value option. During the initial training phase, the options produce Gaussian-distributed rewards (means in parentheses, standard deviation of \$1). (b) All pairwise combinations of options in the transfer test. Blue lines show the originally trained pairs. Red lines show pairs for which absolute and relative values conflict. (c) Prompt designs used in all experiments. The comparisons prompt added explicit comparisons between the local outcomes delivered in each round.}
\end{figure}

The present study uses the machine psychology framework to investigate the ability of LLMs to learn from past outcomes to make reward-maximizing choices. The focus is on relatively simple bandit tasks because of their simplicity and tractability, and because they have shed light on fundamental properties of reward encoding and belief updating in past studies \citep{lefebvre2017behavioural,palminteri2015contextual,palminteri2017confirmation}. In the bandit tasks used here, options are grouped in fixed pairings (or triplets) during the initial training phase. Figure~\ref{Fig1}a shows an example task with four pairs of options (i.e., “contexts”), adapted from \citep{hayes2022reinforcement}. Each pair has a lower value option and a higher value option with Gaussian-distributed rewards. The goal is to make choices that maximize payoffs, which means that agents should learn to exploit the higher value option in each pair. People do this very efficiently with trial-by-trial feedback from both options \citep{hayes2022reinforcement}.

Analyzing human behavior in this kind of bandit task has revealed that outcomes are encoded in a non-normative fashion. That is, people seem to encode rewards in a relative or context-dependent manner, such that the subjective value of an option depends on how its outcomes compare to the outcomes of the other option in the same pair \citep{palminteri2021context}. This is clearly seen by including a transfer test after the training phase, where agents are given choices between all pairwise combinations of options (Figure~\ref{Fig1}b). The objective is still to make reward-maximizing choices, but feedback is no longer provided. If agents learned absolute values, they should be highly accurate at choosing the reward-maximizing options. On the other hand, if they only learned relative values, they should make correct choices whenever the maximizing option has a higher relative value (e.g., in the trained choice pairs; blue lines in Figure~\ref{Fig1}b), but wrong choices whenever the maximizing option has a lower relative value (red lines in Figure~\ref{Fig1}b). People tend to exhibit a relative value bias in the transfer test, frequently failing to select maximizing options that were paired with a better option during training \citep{hayes2022reinforcement}. While relative encoding may facilitate the learning of subjective values within the original training context, it can undermine the ability to generalize learned values to new contexts \citep{bavard2018reference,klein2017learning}. Further, relative value bias in humans is strongly modulated by structural features of the task, including the availability of feedback from nonchosen options \citep{bavard2018reference}, the ordering of the training trials \citep{bavard2021two,hayes2023effects}, and task framing or expectations \citep{hayes2022reinforcement,juechems2022human}.

The present study examines relative outcome encoding in LLMs using simulated bandit tasks. While there is some evidence for relative value bias in LLMs, it is currently limited to a single task and just two models \citep{hayes2024relative}. Here, results are presented from a multi-task, multi-model investigation that encompassed five distinct tasks and four LLMs. Two different prompt designs are examined, one in which the outcomes in each round are listed in a neutral fashion (standard prompt), and another that incorporates explicit comparisons between the outcomes in each round (comparisons prompt) (Figure~\ref{Fig1}c). Explicit comparisons should make it easier to differentiate between options in the same training context. In humans, factors that make the learning task easier tend to increase relative processing \citep{bavard2021two}. Based on this finding and preliminary evidence from a prior study \citep{hayes2024relative}, it is predicted that relative value bias would be observed in both prompt conditions but to a greater degree in the comparisons condition.

To foreshadow the results, choice of reward-maximizing options was generally above chance for all tasks and models. The models were generally able to learn from past outcomes presented in-context and generalize to unseen choice sets. In four out of the five tasks, behavioral signatures of relative value bias were also found. LLM agents tended to choose options with better relative value at rates that exceeded those of an ideal, reward-maximizing agent, especially in the comparisons prompt condition (Figure~\ref{Fig1}c). While the presence of explicit outcome comparisons generally improved training performance, it had an opposite effect on transfer performance. This mirrors a trade-off observed in humans, where factors that favor relative outcome valuation increase performance in the original training contexts but undermine generalization to new contexts \citep{bavard2021two}. Additionally, computational cognitive modeling revealed that LLM behavior is well-described by simple RL algorithms that encode a combination of relative and absolute outcomes. Taken together, these findings have important implications for the study of in-context reinforcement learning in LLMs, highlighting the prevalence of human-like relative value biases.

\section{Method}

\subsection{Models}

Four LLMs were tested in the main experiments. All four are based on the transformer architecture (decoder-only), pretrained on next-token prediction, and fine-tuned for chat or instruction following. Two proprietary models, gpt-3.5-turbo-0125 and gpt-4-0125-preview \citep{achiam2023gpt}, were accessed via the OpenAI API. The other two models, llama-2-70b-chat \citep{touvron2023llama} and mixtral-8x7b-instruct \citep{jiang2024mixtral}, are open-source and were accessed via the Hugging Face Inference API. All models were accessed between March and April 2024. The models’ temperature parameter was set to zero to obtain deterministic responses \citep{coda2024cogbench,schubert2024context}. All other parameters were kept at their default values.

\subsection{Bandit tasks}

Five bandit tasks were selected from prior studies \citep{bavard2018reference, vandendriessche2023contextual, hayes2022reinforcement, bavard2023functional, hayes2023testing} that varied on a range of structural features. Table~\ref{Tab1} summarizes the major differences between them, and Appendix \ref{AppendixA} presents a full summary of the options in each. Henceforth, the abbreviations in Table~\ref{Tab1} will be used to refer to the tasks.

At a broad level, four to ten options were grouped together in fixed training contexts. The training contexts were mostly binary (comprised of two options each), but one task (BP2023) included two ternary contexts. The options produced probabilistic rewards from either a Bernoulli or Gaussian distribution. Two of the training contexts in the B2018 task included losses. For the tasks with Bernoulli-distributed outcomes, the relative frequencies were made to match the underlying probabilities exactly in the simulations. 

All tasks consisted of a training phase and a transfer test. Each training context was presented 12-30 times during training for a total of 48-60 trials, randomly interleaved. The transfer test consisted of one or two (V2023 only) choices for all possible pairwise combinations of options.

\begin{table}
  \caption{Summary of the bandit tasks}
  \label{Tab1}
  \centering
  \begin{tabular}{llllll}  
    \toprule

                           & Task \#1   & Task \#2   & Task \#3   & Task \#4     & Task \#5   \\
                           & (B2018)    & (V2023)    & (HW2023a)  & (BP2023)     & (HW2023b)  \\
    \midrule
    Options                & 8          & 4          & 8          & 10           & 8          \\
    Training contexts (\#) & binary (4) & binary (2) & binary (4) & binary (2),  & binary (4) \\         
                           &            &            &            & ternary (2)  &            \\
    Reward distribution    & Bernoulli  & Bernoulli  & Gaussian   & Gaussian     & Bernoulli  \\
    Reward range           & [-1, 1]    & [0, 1]     & $\sim$[12, 39] & $\sim$[8, 92] & [10, 44]  \\
    Reward currency        & euros      & points     & dollars    & dollars      & dollars    \\
    Training trials*       & 12         & 30         & 15         & 15           & 15         \\
    Transfer trials        & 28         & 12         & 28         & 45           & 28         \\
    Adapted from           & \citep{bavard2018reference} & \citep{vandendriessche2023contextual} & \citep{hayes2022reinforcement} & \citep{bavard2023functional} & \citep{hayes2023testing} \\
    \bottomrule
    \multicolumn{6}{p{5in}}{*Training trials are given per training context, so to infer the total number of trials in the training phase this value should be multiplied by the value in ``Training contexts (\#).''}
  \end{tabular}
\end{table}

\subsection{Experimental procedures}

The experimental procedures were based on prior studies \citep{binz2023using, coda2024cogbench, hayes2024relative, schubert2024context}. All instructions, choice stimuli, and rewards were translated to natural language. The options were described as ``slot machines,'' with the letters A-J randomly assigned at the beginning of each simulation run depending on the number of options in the task. The instructions varied slightly across tasks, but the gist was always that the agent would be making several choices with the goal of maximizing payoffs (see Appendix \ref{AppendixB} for task-specific instructions). 

In each round the task, the end of the prompt stated, ``You now face a choice between slot machine [X] and slot machine [Y]'' with X and Y replaced by the letters that were assigned to the currently available options. The order in which the options were listed was randomized on a trial-by-trial basis. After restating the goal to maximize payoffs, the last sentence specified the desired response format: ``Which slot machine do you choose? Give your answer like this: I would choose slot machine \_. Do not explain why.'' This format was chosen to facilitate data analysis and avoid lengthy responses.     

After the LLM responded with the chosen slot machine, if the current trial was part of the training phase, outcomes from the chosen and nonchosen option(s) were drawn from the corresponding distributions and appended to the history of previous outcomes (complete feedback). The updated outcome history was then used in the prompt for the next trial. If the current trial was part of the transfer test, the outcome history was left unchanged (no feedback). Two prompt designs were tested. In the standard prompt, the outcomes for both options in each round were listed in a neutral fashion. In the comparisons prompt, explicit comparisons between the outcomes were added (see Figure \ref{Fig1}c). All other parts of the prompt were the same in both conditions.

30 independent experiments were run for each combination of task (5 levels), LLM (4 levels), and prompt condition (2 levels), for a total of 1200 runs. All experiments were carried out on a CPU.

\section{Results}

\subsection{Choice accuracy}

The explicit goal in each bandit task was to maximize payoffs. Choice accuracy, defined as the proportion of reward-maximizing choices, was computed for each experiment run and analyzed using a 5 (Task) $\times$ 4 (LLM) $\times$ 2 (Prompt) between-subjects analysis of variance (ANOVA).\footnote{Prior to analysis, \textit{n} = 1401 invalid trial-level responses that did not conform to the specified format were removed from the data (1.36\% of all choice trials). Most of these (\textit{n} = 1204) occurred in the BP2023 task.} Means and margins of error (95\% confidence level) are reported for key comparisons. The training phase and transfer test were analyzed separately. 

\begin{figure}[ht]
  \centering
  \includegraphics[width=5.5in]{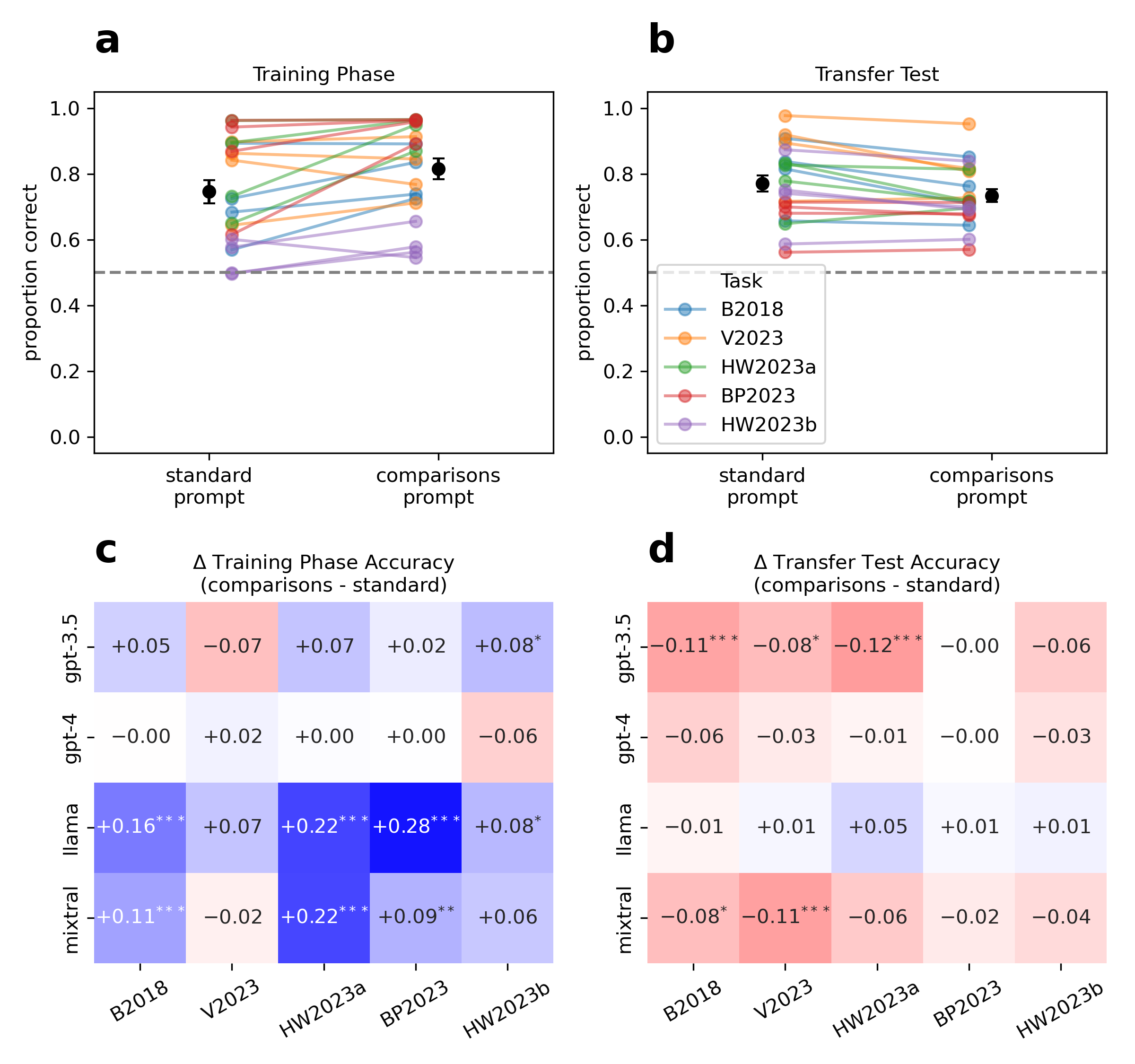}
  \caption{(a-b) Mean choice accuracy (proportion of reward-maximizing choices) in the training phase and transfer test. Each colored point represents the mean accuracy for a specific combination of task, model, and prompt design across 30 runs (lines connect the same task/model combination). Means and standard errors are also shown. (c-d) Pairwise contrasts for the effect of prompt design, broken down by task and model. *\textit{p} < .05 **\textit{p} < .01 ***\textit{p} < .001 (Bonferroni-adjusted for 20 tests).}
  \label{Fig2}
\end{figure}

Training accuracy was generally well above chance, indicating successful in-context learning (Figure \ref{Fig2}a). Averaging across tasks and models, training accuracy was significantly higher using the comparisons prompt (.815 $\pm$ .008) compared to the standard prompt (.746 $\pm$ .008), \textit{F}(1, 1160) = 147.39, \textit{p} < .001, $\eta_p^2$ = 0.11. All other main effects and interactions were also significant (\textit{p}s < .001; see Appendix \ref{AppendixC} for full ANOVA results). Of note, training accuracy tended to be higher for the GPT models, especially gpt-4-0125-preview, and for the tasks with Gaussian reward distributions (HW2023a, BP2023).

Average transfer accuracy was also above chance (Figure \ref{Fig2}b).\footnote{Choices between options with equal expected value were excluded from this analysis. These only occurred in the BP2023 task.} However, unlike in training, transfer accuracy was significantly lower, on average, using the comparisons prompt (.734 $\pm$ .007) compared to the standard prompt (.771 $\pm$ .007), \textit{F}(1, 1160) = 53.52, \textit{p} < .001, $\eta_p^2$ = 0.04. The other main effects and two-way interactions were significant (\textit{p}s < .005), but the three-way interaction was not (Appendix \ref{AppendixC}). The highest transfer accuracy was achieved by gpt-4-0125-preview (.847 $\pm$ .01), and the lowest by llama-2-70b-chat (.641 $\pm$ .01). Transfer accuracy also varied across tasks, being highest in the V2023 task (.852 $\pm$ .011) and lowest in the BP2023 task (.662 $\pm$ .011).

The bottom panels of Figure \ref{Fig2} show the effect of prompt design for all combinations of task and model. In the training phase, the comparisons prompt was most beneficial for the two open-source models, especially in the tasks with Gaussian reward distributions (Figure \ref{Fig2}c). In contrast, gpt-4-0125-preview, which performed at or near ceiling by the end of training in four of the five tasks, did not benefit from explicit comparisons. The pattern was considerably different for the transfer test, where the comparisons prompt had a mostly negative effect (Figure \ref{Fig2}d). The one exception was llama-2-70b-chat, for which explicit comparisons led to a slight overall improvement in transfer accuracy.

\subsection{Relative value bias}

To analyze relative value bias, we need a definition of ``relative value'' that works for all five bandit tasks. One could define the relative value of an option by whether it was the optimal choice (based on expected value) in its training context. However, people sometimes prefer sub-optimal options if they give better outcomes \textit{most of the time} \citep{hayes2023testing}. Strikingly, LLMs also exhibited this tendency in the HW2023b task (see Figure \ref{Fig16}). Thus, we define the relative value of an option based on how most of its outcomes compare to the outcomes of the other option(s) in the training context. Options that frequently give better local outcomes have higher relative value, and options that frequently give worse local outcomes have lower relative value.

Relative value bias was analyzed using the models’ choices in the transfer test. On some trials, the option with higher relative value (from the training phase) was the reward-maximizing choice, but on other trials, the option with higher relative value was the sub-optimal choice (see Figure \ref{Fig1}b for an example).\footnote{Choices between options with tied relative values were excluded from this analysis.} One can readily calculate the rate at which an ideal, reward-maximizing agent would end up choosing options with higher relative value. The ideal choice rate differed across the five tasks, ranging from .50 in HW2023b to .969 in BP2023. If the empirical choice rate exceeded the ideal, this was taken as an indicator of relative value bias.

LLMs frequently chose options with higher relative value in excess of an ideal agent, mainly when using the comparisons prompt, and most noticeably in the B2018, HW2023a, and HW2023b tasks (Appendix \ref{AppendixD}). A Task $\times$ LLM $\times$ Prompt ANOVA confirmed that options with higher relative value were chosen at a higher rate with the comparisons prompt (.829 $\pm$ .009) than with the standard prompt (.698 $\pm$ .009), \textit{F}(1, 1159) = 440.88, \textit{p} < .001, $\eta_p^2$ = 0.28, averaging across tasks and models. All other main effects and interactions were significant as well (\textit{p}s < .001). Table \ref{Tab2} shows the estimated marginal mean choice rates as a function of prompt design and task, collapsing across LLMs. When compared to an ideal agent, the comparisons prompt led to significant relative value bias in four of the five tasks, while the standard prompt led to significant relative bias in two tasks.

\begin{table}
    \caption{Estimated mean choice rates for options with higher relative value in the transfer test.}
    \label{Tab2}
    \centering
    \begin{tabular}{llcccc}
      \toprule
         Prompt      & Task    & Ideal & Mean  & 95\% CI        & Bias        \\
      \midrule
         Standard    & B2018   & 0.625 & 0.643 & [0.623, 0.662] &             \\
                     & V2023   & 0.750 & 0.738 & [0.718, 0.757] &             \\
                     & HW2023a & 0.625 & 0.721 & [0.701, 0.740] & \checkmark  \\
                     & BP2023  & 0.969 & 0.811 & [0.792, 0.831] &             \\
                     & HW2023b & 0.500 & 0.577 & [0.558, 0.597] & \checkmark  \\
         Comparisons & B2018   & 0.625 & 0.776 & [0.757, 0.795] & \checkmark  \\
                     & V2023   & 0.750 & 0.787 & [0.768, 0.807] & \checkmark  \\
                     & HW2023a & 0.625 & 0.926 & [0.906, 0.945] & \checkmark  \\
                     & BP2023  & 0.969 & 0.936 & [0.916, 0.955] &             \\
                     & HW2023b & 0.500 & 0.721 & [0.701, 0.740] & \checkmark  \\
      \bottomrule
      \multicolumn{6}{p{4in}}{Note: Results are collapsed across LLMs. Bias column indicates whether there was a significant relative value bias (if the 95\% CI for the mean choice rate was above the ideal agent).}
    \end{tabular}
\end{table}

\subsection{Computational cognitive modeling}

Computational cognitive models were used to provide an interpretable, mechanistic account of LLM in-context learning in the bandit tasks. The models were simple RL algorithms with three basic components: (1) a subjective value function, which determines how rewards are encoded, (2) a learning function, which determines how expectancies are updated in response to feedback, and (3) a response function, which maps expectancies onto actions. 

The general form of the subjective value function is as follows:
\begin{equation}
    v(x_{i,t}) = (1 - \omega) \cdot x_{i,t}^{ABS} + \omega \cdot x_{i,t}^{REL}
\label{Eq1}
\end{equation}
where $x_{i,t}$ is the \textit{i}th reward on trial \textit{t}, $x_{i,t}^{ABS}$ is the value of $x_{i,t}$ normalized by the full range of rewards experienced so far across all training contexts,\footnote{Using the range normalization formula: $(x - min)/(max - min)$.} $x_{i,t}^{REL}$ is the value of $x_{i,t}$ normalized by the range of rewards experienced on the current trial only, and $\omega$ is the relative encoding parameter. Similar weighted encoding schemes are frequently used to model human RL behavior \citep{bavard2018reference, hayes2021regret, hayes2023testing, molinaro2023intrinsic}. Note that if $\omega = 0$, subjective values are proportional to the absolute reward magnitudes and the model predicts no effects of the local training context. We refer to this special case as the ABS model. On the other hand, if $\omega = 1$, the subjective value of an outcome is entirely determined by how it compares to the other outcome(s) from the local training context experienced on that trial. We refer to the more general, hybrid model with $\omega$ freely estimated as the REL model.

Once subjective values are formed, learning is modeled using a basic prediction error-driven updating function \citep{rescorla1972theory}:
\begin{equation}
    Q_{t+1}(a_i) = Q_{t}(a_i) + \alpha \cdot (v(x_{i,t}) - Q_{t}(a_i))
\label{Eq2}
\end{equation}
where $Q_{t+1}(a_i)$ is the updated expectancy for option $a_i$, $Q_{t}(a_i)$ is the prior expectancy, and $\alpha$ is the learning rate parameter. In bandit tasks, LLMs seem to update expectancies more in response to feedback that confirms prior beliefs and discount feedback that contradicts them \citep{schubert2024context}. This propensity has also been observed in humans and is thought to represent a kind of confirmation bias \citep{palminteri2017confirmation}. Based on that work, we include models with separate learning rates for confirmatory outcomes ($\alpha_{CON}$) and disconfirmatory outcomes ($\alpha_{DIS}$).\footnote{Confirmatory outcomes are when the chosen option's outcome is better than expected (positive prediction error) or the unchosen option's outcome is worse than expected (negative prediction error). Disconfirmatory outcomes are the reverse.}

Learned expectancies are mapped to choice probabilities using the softmax function with an inverse temperature parameter $\beta$: 
\begin{equation}
    p(a_i) = \frac{\exp(\beta \cdot Q_t(a_i) + b \cdot \delta(a_i))}{\sum_{k} \exp(\beta \cdot Q_t(a_k) + b \cdot \delta(a_k))}
\label{Eq3}
\end{equation}
where $\delta(\cdot)$ is an indicator function that equals 1 if its argument is the first option listed in the prompt. Positive values of $b$ increase preference for the first option, regardless of its value. Thus, choice probabilities depend on both learned expectancies and, potentially, response bias. Finally, some models used separate inverse temperatures ($\beta$) for the training phase and transfer test.

The factorial combination of absolute or (partially) relative encoding, one or two learning rates, and one or two inverse temperatures results in eight candidate models. Models were fit to the choice data using maximum likelihood methods. A single set of parameters was estimated for each combination of LLM, task, and prompt, each time pooling the data across the 30 experiment runs, with $Q$ values reset to 0.5 on the first trial of each run. Models were compared using the Bayesian information criterion (BIC), where models with lower BIC are preferred.

Figure \ref{Fig3} shows the best-fitting models across tasks, prompt designs, and LLMs (see Appendix \ref{AppendixE} for the BIC values in each task). With the exception of the V2023 task, models that incorporated relative encoding were more frequently selected as the best description of LLM behavior. Models that used separate learning rates and/or separate inverse temperatures were more frequently selected than the models with a single learning rate or inverse temperature. 

\begin{figure}[ht]
  \centering
  \includegraphics[width=5.5in]{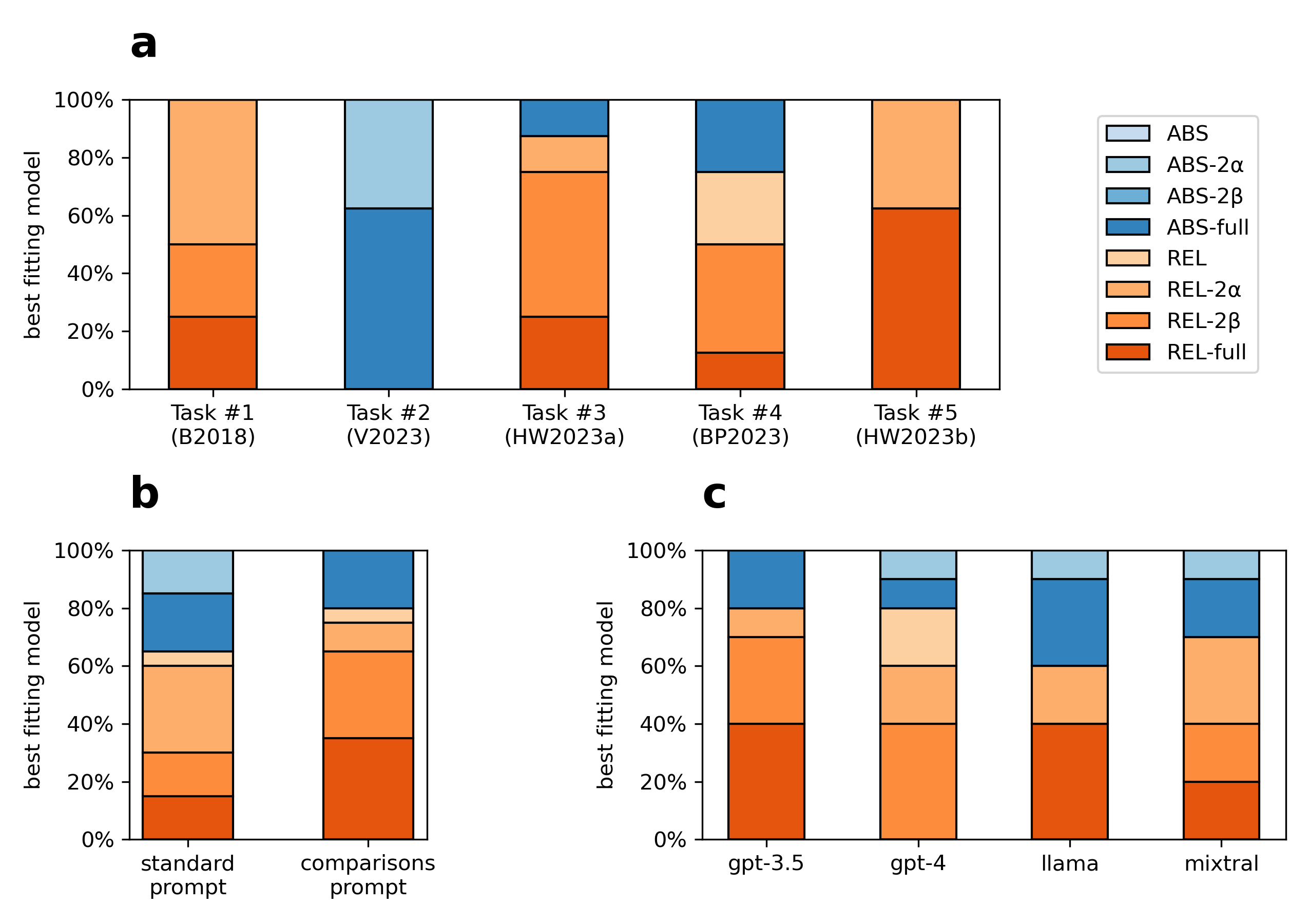}
  \caption{Best-fitting models (\%) across tasks (a), prompt designs (b), and LLMs (c).}
  \label{Fig3}
\end{figure}

Appendix \ref{AppendixD} shows the fit of the ABS-full and REL-full models to the data. The ABS-full model was able to capture some of the relative value biases even though it assumes unbiased outcome encoding. This was likely due to the use of separate learning rates: If $a_{CON} > a_{DIS}$, the models tend to repeat previous choices, which would generally favor options with higher relative value. However, it was clearly not able to capture the magnitude of the biases as well as the REL-full model. See Appendix \ref{AppendixF} for plots showing the fit of the REL-full model to the training and transfer patterns in each task.

\begin{figure}[ht]
  \centering
  \includegraphics[width=5.5in]{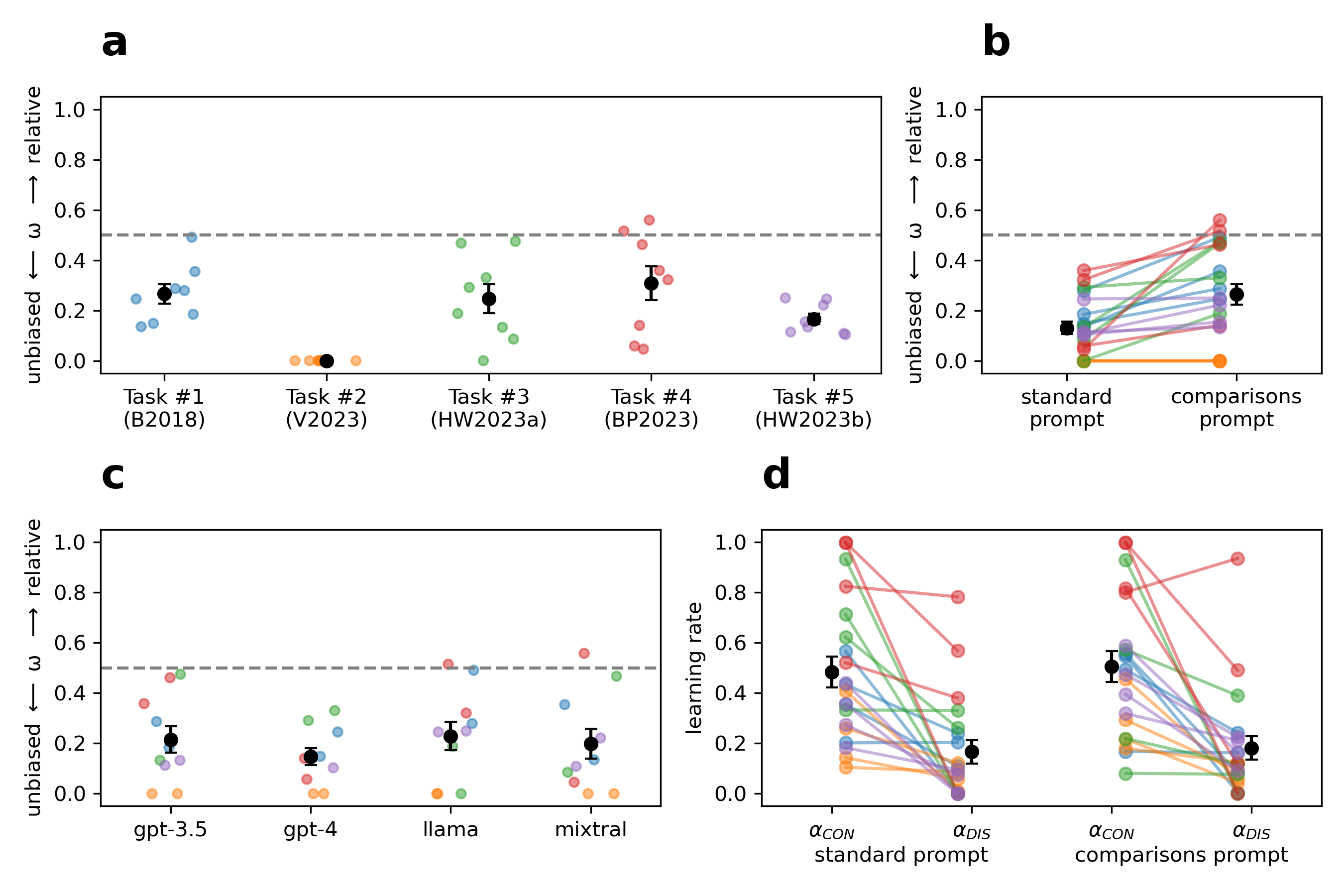}
  \caption{(a) Estimated relative encoding parameters across tasks, (b) prompt designs, and (c) LLMs. (d) Estimated learning rates. In each panel, black points show the means and standard errors.}
  \label{Fig4}
\end{figure}

The parameters of the REL-full model were examined next. The mean estimates for the relative encoding parameter ($\omega$) were consistently above 0 for all tasks except V2023 (Figure \ref{Fig4}a). In line with the model-free results, $\omega$ estimates were significantly higher with the comparisons prompt (.264 $\pm$ .088) than with the standard prompt (.131 $\pm$ .054), \textit{t}(19) = 4.14, \textit{p} < .001, \textit{d} = 0.93 (paired \textit{t}-test; Figure \ref{Fig4}b). All four LLMs exhibited similar degrees of relative encoding on average (Figure \ref{Fig4}c). Finally, consistent with prior work \citep{schubert2024context}, the estimates for $\alpha_{CON}$ were higher than the estimates for $\alpha_{DIS}$ using both the standard prompt (.483 $\pm$ .132 vs. .166 $\pm$ .099) and the comparisons prompt (.505 $\pm$ .133 vs. .182 $\pm$ .102) (Figure \ref{Fig4}d). 

Parameters that control the mapping from expectancies to choice probabilities were also examined. The average inverse temperature ($\beta$) was higher in training (20.243 $\pm$ 8.097) than in transfer (11.265 $\pm$ 3.351), which may be interpreted as a kind of ``forgetting,'' and the average position bias ($b$) was significantly positive (1.176 $\pm$ 0.145), indicating a bias toward choosing the first option listed. See Appendix \ref{AppendixG} for a comparison of fitted parameters across LLMs.

\subsection{Analysis of hidden states}

The four LLMs that were tested are all fine-tuned for chat or instruction following. In an additional set of experiments (Appendix \ref{AppendixH}), we ran gemma-7b \citep{team2024gemma}, a pretrained model with no fine-tuning, through the HW2023a task and found that it too showed robust relative value biases. In the transfer test, the model selected options with higher relative value 61\% of the time using the standard prompt, but 91\% of the time using the comparisons prompt (ideal agent: 62.5\%). This preliminary evidence suggests that the observed biases are not entirely due to fine-tuning procedures, and can emerge from pretraining on next-token prediction alone.

Interestingly, the effect of prompt design was strongly reflected in the model's internal states. To demonstrate, we ran gemma-7b through the transfer test \textit{n}~= 100 times and extracted the final hidden layer activations for the last token in the choice prompt on each trial (see Appendix \ref{AppendixH} for details). A linear regression model was then fit to the activations in each hidden unit separately (\textit{n}~= 3072 regressions). The two predictors were the absolute and relative value differences between the available options on each trial. Using an extremely conservative correction for multiple tests, the slope for the relative value predictor was significant for only 3\% of the hidden units using the standard prompt, but 72\% using the comparisons prompt. In contrast, the absolute value predictor was significant for 19\% of the hidden units with the standard prompt but just 2\% with the comparisons prompt. Thus, the comparisons prompt enhanced the representation of relative value information and suppressed the representation of absolute value information in the model's hidden layers.

\section{Conclusions}

The present research examined in-context learning in LLMs performing bandit tasks \citep{binz2023using, coda2024cogbench, hayes2024relative, schubert2024context}. In addition to a previously-established confirmation bias \citep{schubert2024context}, we find that LLMs exhibit a relative value bias that can undermine their ability to transfer learned action values to new choice problems \citep{bavard2018reference,bavard2021two,hayes2022reinforcement,klein2017learning,palminteri2021context}. LLM behavior was well-approximated by a simple RL algorithm that combines absolute and relative value signals at the outcome encoding stage. The finding that LLMs are biased reinforcement learners is important for real-world applications of LLMs as decision-making agents.

A limitation of the current work is its focus on models that have been fine-tuned for chat or instruction following using techniques such as reinforcement learning from human feedback \citep{christiano2017deep} or direct preference optimization \citep{rafailov2024direct}. However, we present preliminary evidence that the biases we observed can appear even in the absence of fine-tuning and human feedback. Another limitation is that we did not consider prompting strategies that may reduce relative value bias. In prior work, instructing the models to estimate expected payoffs before making a choice all but eliminated the bias \citep{hayes2024relative}. Other strategies, such as zero-shot chain of thought prompting \citep{kojima2022large}, could prove effective. Future work should continue searching for prompting strategies that can mitigate decision-making biases while extending this line of research to a broader set of LLMs and task designs \citep{coda2024cogbench}.

\begin{ack}
The authors thank Can Demircan for helpful discussions regarding the analysis of hidden unit activations.
\end{ack}

\section*{Data and Code Availability}
Data and code are available in the GitHub repository \url{https://github.com/william-hayes/LLMs-biased-RL}.

\bibliography{biblio}


\newpage

\appendix

\section{Overview of bandit tasks}
\label{AppendixA}

In each task, the options were grouped into separate training contexts, each containing a lower value option (L) and a higher value option (H). In the BP2023 task only, there were two ternary contexts containing a third, medium value option (M). Each option produced rewards from a Bernoulli (B2018, V2023, HW2023b) or Gaussian distribution (HW2023a, BP2023).

\begin{figure}[h]
  \centering
  \includegraphics[width=5.5in]{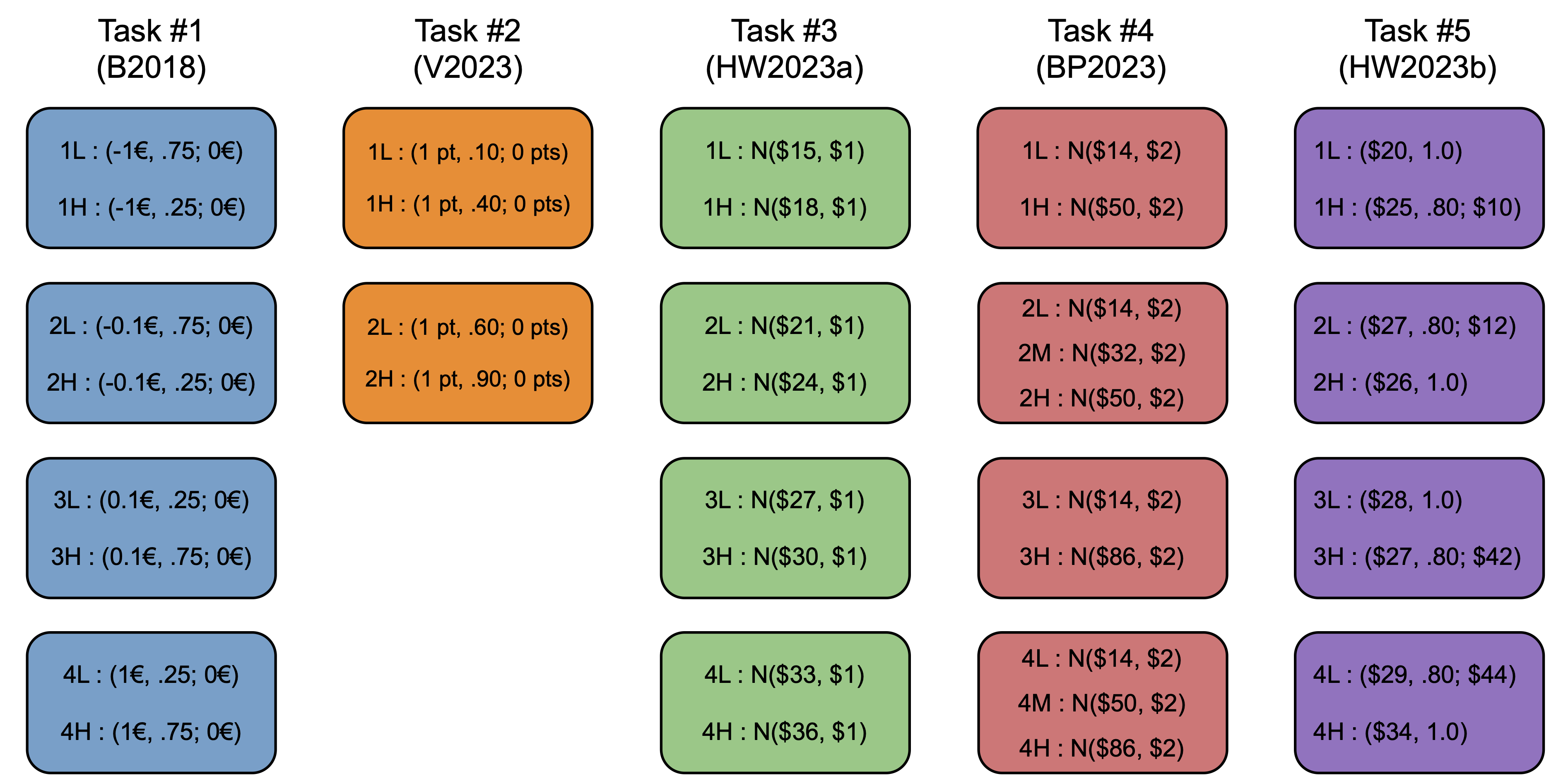}
  \caption{A list of the options in each bandit task. The notation $(x, p; y)$ means that $x$ occurred with probability $p$, otherwise $y$. The notation $N(a, b)$ refers to a normal distribution with mean $a$ and standard deviation $b$.}
  \label{Fig5}
\end{figure}

\clearpage

\section{Task-specific instructions}
\label{AppendixB}

The instructions differed slightly across the five bandit tasks. Below are the exact instructions that appeared at the beginning of the prompts. 

\begin{tcolorbox}[colback=blue!5!white,colframe=blue!75!black,title=Task \#1 (B2018)]
The aim of this task is to maximize your payoffs.\\
There are several slot machines that deliver wins and losses with different probabilities.\\
On each round, you will be asked which of two slot machines you wish to play.\\
Seeking monetary rewards and avoiding monetary losses are equally important.\\
Your total payoff will be the cumulative sum of the money you win across all rounds of the game.
\end{tcolorbox}

\begin{tcolorbox}[colback=blue!5!white,colframe=blue!75!black,title=Task \#2 (V2023)]
You are playing a game that involves choosing between different slot machines.\\
Each slot machine gives 1 point with a particular probability, otherwise 0 points.\\
Some slot machines have a higher probability of reward than others.\\
The goal is to maximize your total payoff over the course of several rounds.\\
Your total payoff will be the cumulative sum of the points you win across all rounds of the game.
\end{tcolorbox}

\begin{tcolorbox}[colback=blue!5!white,colframe=blue!75!black,title=Task \#3 (HW2023a)]
You are playing a game with the goal of winning as much money as possible over the course of several rounds.\\
In each round, you will be asked which of two slot machines you wish to play.\\
Some slot machines win more money than others on average.\\
Your total payoff will be the cumulative sum of the money you win across all rounds of the game.\\
Remember that your goal is to maximize your total payoff.
\end{tcolorbox}

\begin{tcolorbox}[colback=blue!5!white,colframe=blue!75!black,title=Task \#4 (BP2023)]
In this task, you will be given information about several slot machines in order to decide which ones you want to play.\\
Some slot machines win more money than others on average.\\
On each trial, you will be asked to choose between two or three different slot machines.\\
Your goal is to make choices that maximize your total payoffs. In other words, you should try to win as much money as possible.\\
Your total payoff will be the cumulative sum of the money you win across all rounds of the game.
\end{tcolorbox}

\begin{tcolorbox}[colback=blue!5!white,colframe=blue!75!black,title=Task \#5 (HW2023b)]
In this task, you will be given information about several slot machines in order to decide which ones you want to play.\\
Some slot machines win more money than others on average.\\
Your goal is to make choices that maximize your total payoffs. In other words, you should try to win as much money as possible.\\
Your total payoff will be the cumulative sum of the money you win across all rounds of the game.
\end{tcolorbox}

\clearpage

\section{Choice accuracy: Additional results}
\label{AppendixC}

Below are the results from the ANOVAs on choice accuracy (i.e., proportion of reward-maximizing choices) in the training phase and transfer test. Primary focus was on the main effect of Prompt, which represents the average effect of prompt design across tasks and LLMs. 

\begin{table}[h]
    \centering
    \caption{ANOVA results for training phase accuracy.}
    \begin{tabular}{lllllll}
        \toprule
         Source & $df$ & $SS$ & $MS$ & $F$ & $p$ & $\eta_p^2$  \\
        \midrule
         Task & 4 & 16.858 & 4.214 &	434.899 &	< .001 &	0.60 \\
         LLM  & 3 &	5.376 &	1.792 &	184.933 &	< .001 &	0.32\\
         Prompt & 1 &	1.428 &	1.428 &	147.391 &	< .001 &	0.11 \\
         Task $\times$ LLM & 12 &	1.832 &	0.153 &	15.755 &	< .001 &	0.14 \\
         Task $\times$ Prompt & 4 &	0.603 &	0.151 &	15.553 &	< .001 &	0.05 \\
         LLM $\times$ Prompt & 3 &	1.222 &	0.407 &	42.039 &	< .001 &	0.10 \\
         Task $\times$ LLM $\times$ Prompt & 12 &	0.582 &	0.048 &	5.004 &	< .001 &	0.05 \\
         Error & 1160 &	11.241 &	0.010 \\
        \bottomrule
    \end{tabular}
    \label{tab:Tab3}
\end{table}

\begin{table}[h]
    \centering
    \caption{ANOVA results for transfer test accuracy.}
    \begin{tabular}{lllllll}
        \toprule
         Source & $df$ & $SS$ & $MS$ & $F$ & $p$ & $\eta_p^2$  \\
        \midrule
         Task & 4 &	4.652 &	1.163 &	152.330 &	< .001 &	0.34 \\
         LLM  & 3 &	6.470 &	2.157 &	282.490 &	< .001 &	0.42\\
         Prompt & 1 &	0.409 &	0.409 &	53.516 &	< .001 &	0.04 \\
         Task $\times$ LLM & 12 &	0.495 &	0.041 &	5.398 &	< .001 &	0.05 \\
         Task $\times$ Prompt & 4 &	0.116 &	0.029 &	3.795 &	.005 &	0.01 \\
         LLM $\times$ Prompt & 3 &	0.345 &	0.115 &	15.054 &	< .001 &	0.04 \\
         Task $\times$ LLM $\times$ Prompt & 12 &	0.127 &	0.011 &	1.392 &	.163 &	0.01 \\
         Error & 1160 &	8.856 &	0.008 \\
        \bottomrule
    \end{tabular}
    \label{tab:Tab4}
\end{table}

\clearpage

\section{Relative value bias: Additional results}
\label{AppendixD}

Below are results from the analysis of transfer test trials where one option had a higher relative value than the other. An option's relative value was determined by the frequency with which it gave better outcomes than the other option(s) in the same local training context. For each task, we can compute the proportion of times an ideal, reward-maximizing agent would choose the option with higher relative value in the transfer test. If the empirical choice rate from an LLM exceeds the ideal choice rate, this is taken as an indicator of relative value bias. 

\begin{figure}[h]
  \centering
  \includegraphics[width=5.5in]{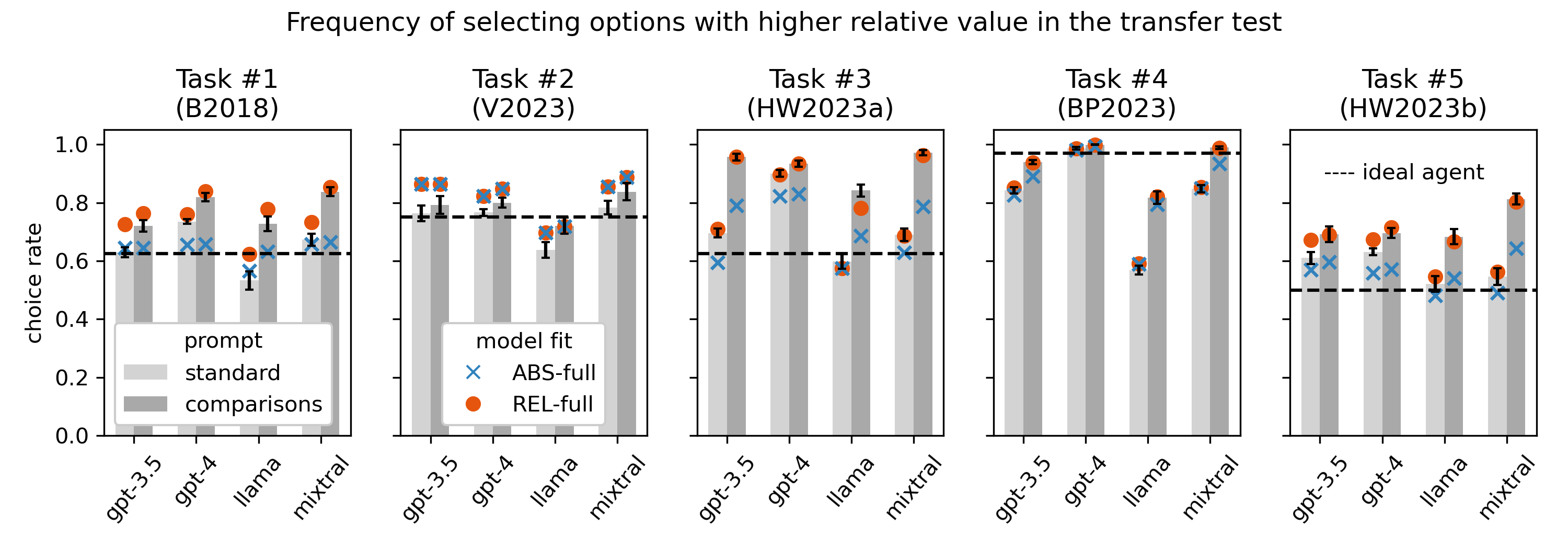}
  \caption{Choice rates for the options with higher relative value in the transfer test. Bars show the empirical data (+/- standard error). Dashed lines show the behavior of an ideal, reward-maximizing agent. The blue X symbols show the fit of the ABS-full model, which assumes unbiased outcome encoding. The orange O symbols show the fit of the REL-full model, which incorporates relative outcome encoding. Both models allow for separate learning rates and inverse temperature parameters. Though the ABS-full model is able to capture some of the relative value biases, the REL-full model fits the data better.}
  \label{Fig6}
\end{figure}

\begin{table}[h]
    \centering
    \caption{ANOVA results for the choice rates of options with higher relative value.}
    \begin{tabular}{lllllll}
        \toprule
         Source & $df$ & $SS$ & $MS$ & $F$ & $p$ & $\eta_p^2$  \\
        \midrule
         Task & 4 &	7.611 &	1.903 &	162.411 &	< .001 &	0.36 \\
         LLM  & 3 &	4.492 &	1.497 &	127.804 &	< .001 &	0.25\\
         Prompt & 1 &	5.165 &	5.165 &	440.882 &	< .001 &	0.28 \\
         Task $\times$ LLM & 12 &	1.271 &	0.106 &	9.038 &	< .001 &	0.09 \\
         Task $\times$ Prompt & 4 &	0.732 &	0.183 &	15.621 & < .001 &	0.05 \\
         LLM $\times$ Prompt & 3 &	0.996 &	0.332 &	28.338 &	< .001 &	0.07 \\
         Task $\times$ LLM $\times$ Prompt & 12 &	0.577 &	0.048 &	4.106 &	< .001 &	0.04 \\
         Error & 1159 &	13.578 &	0.012 \\
        \bottomrule
        \multicolumn{7}{p{4.3in}}{Note: One observation was excluded due to missing data after removing trials with tied relative values.}
    \end{tabular}
    \label{tab:Tab5}
\end{table}

\clearpage

\section{Model comparison by task}
\label{AppendixE}

ABS = absolute encoding, REL = (partially) relative encoding, 2$\alpha$ = separate learning rates, 2$\beta$ = separate inverse temperatures, full = separate learning rates and inverse temperatures.

\begin{figure}[h]
  \centering
  \includegraphics[width=5in]{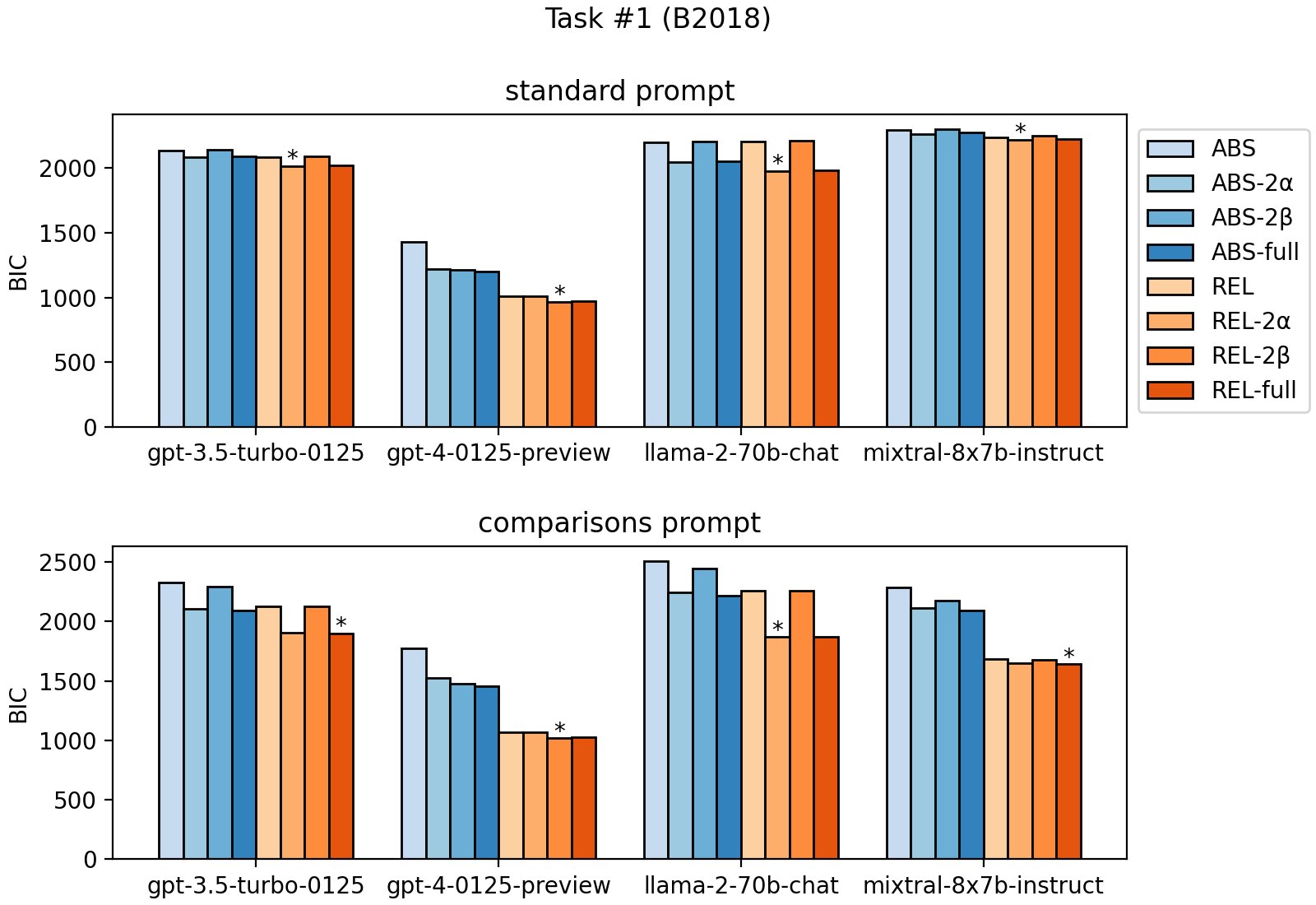}
  \caption{Model BICs in the B2018 task. Asterisks (*) designate the best-fitting models.}
  \label{Fig7}
\end{figure}

\begin{figure}[h]
  \centering
  \includegraphics[width=5in]{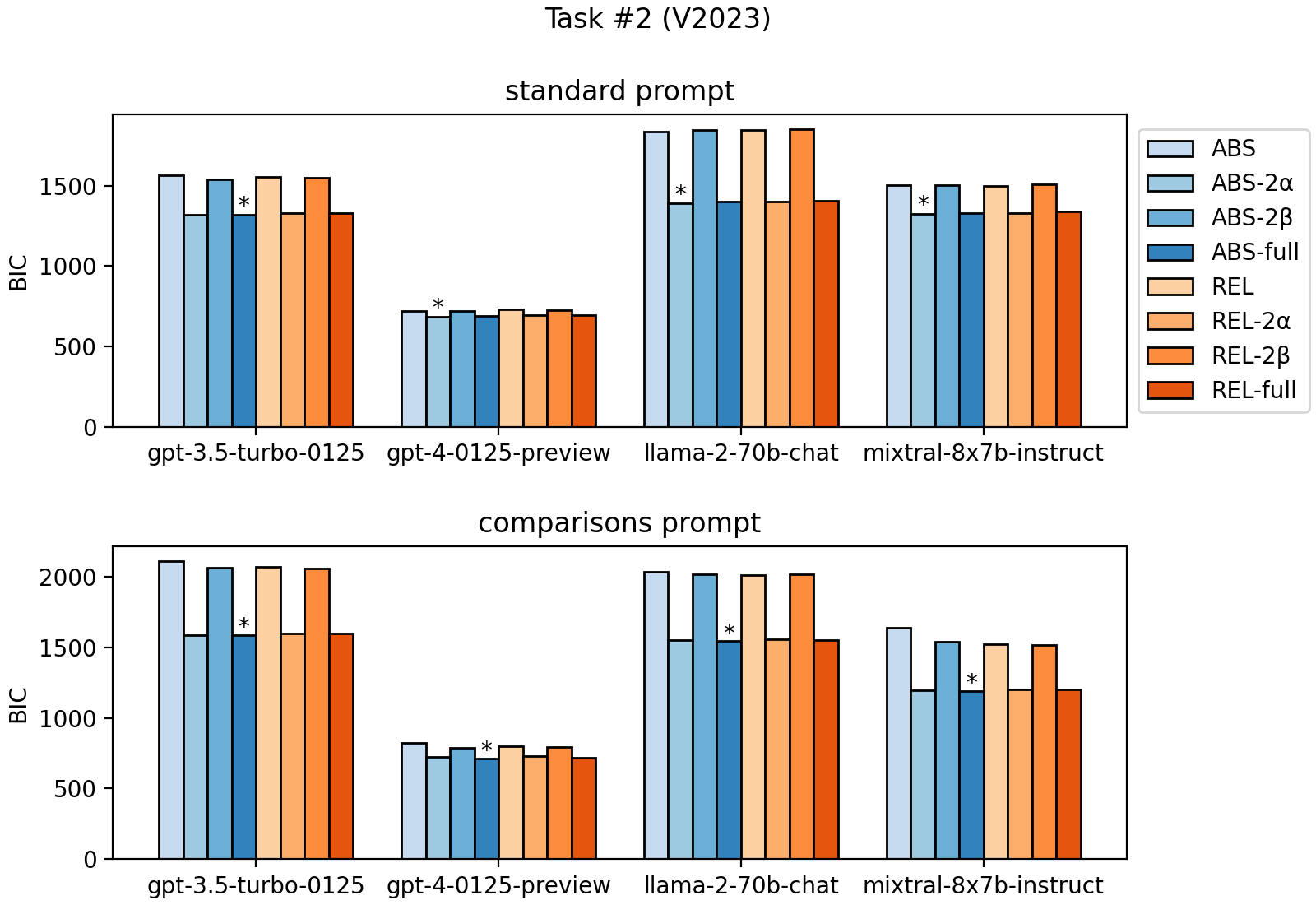}
  \caption{Model BICs in the V2023 task. Asterisks (*) designate the best-fitting models.}
  \label{Fig8}
\end{figure}

\begin{figure}[h]
  \centering
  \includegraphics[width=5in]{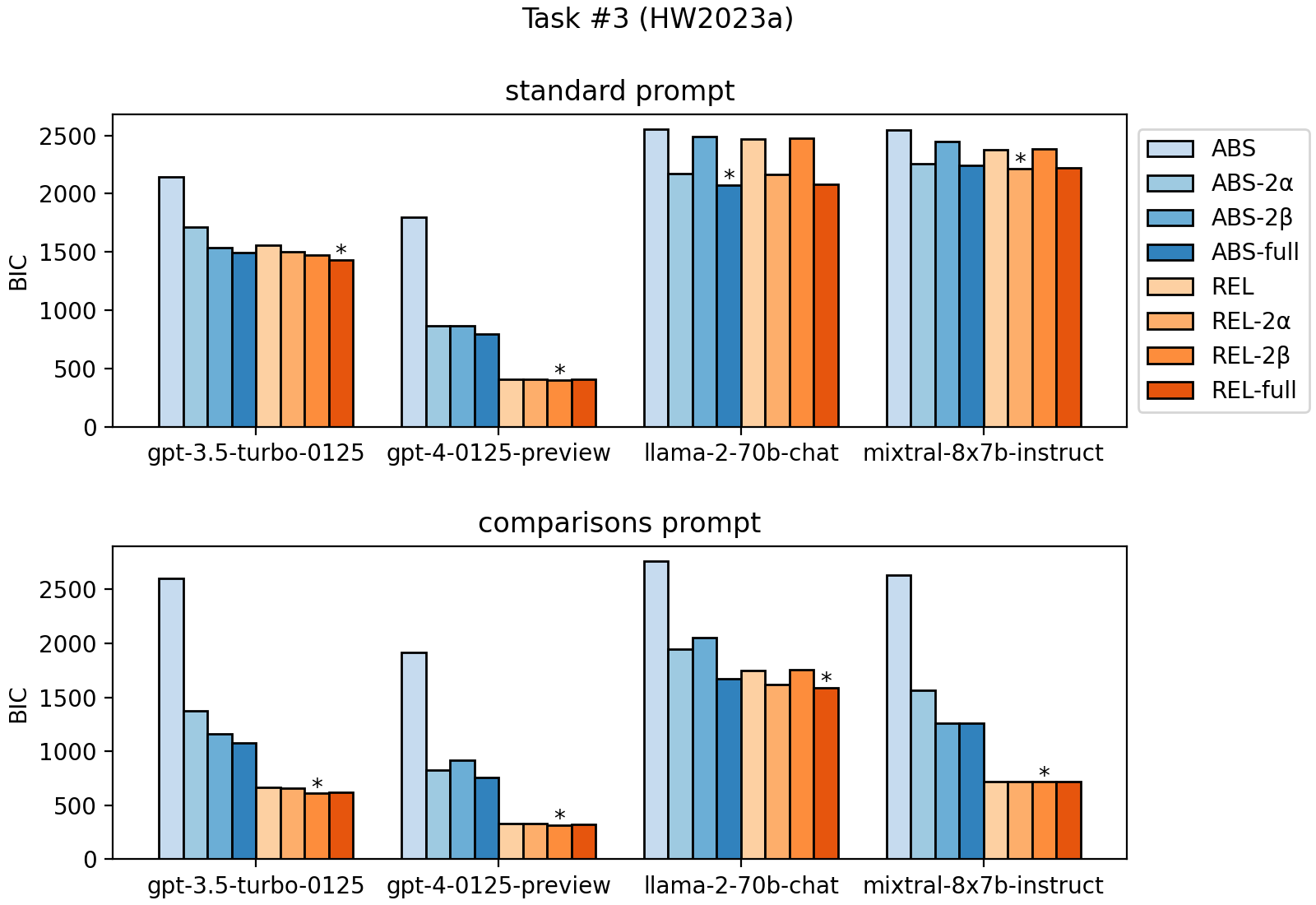}
  \caption{Model BICs in the HW2023a task. Asterisks (*) designate the best-fitting models.}
  \label{Fig9}
\end{figure}

\begin{figure}[h]
  \centering
  \includegraphics[width=5in]{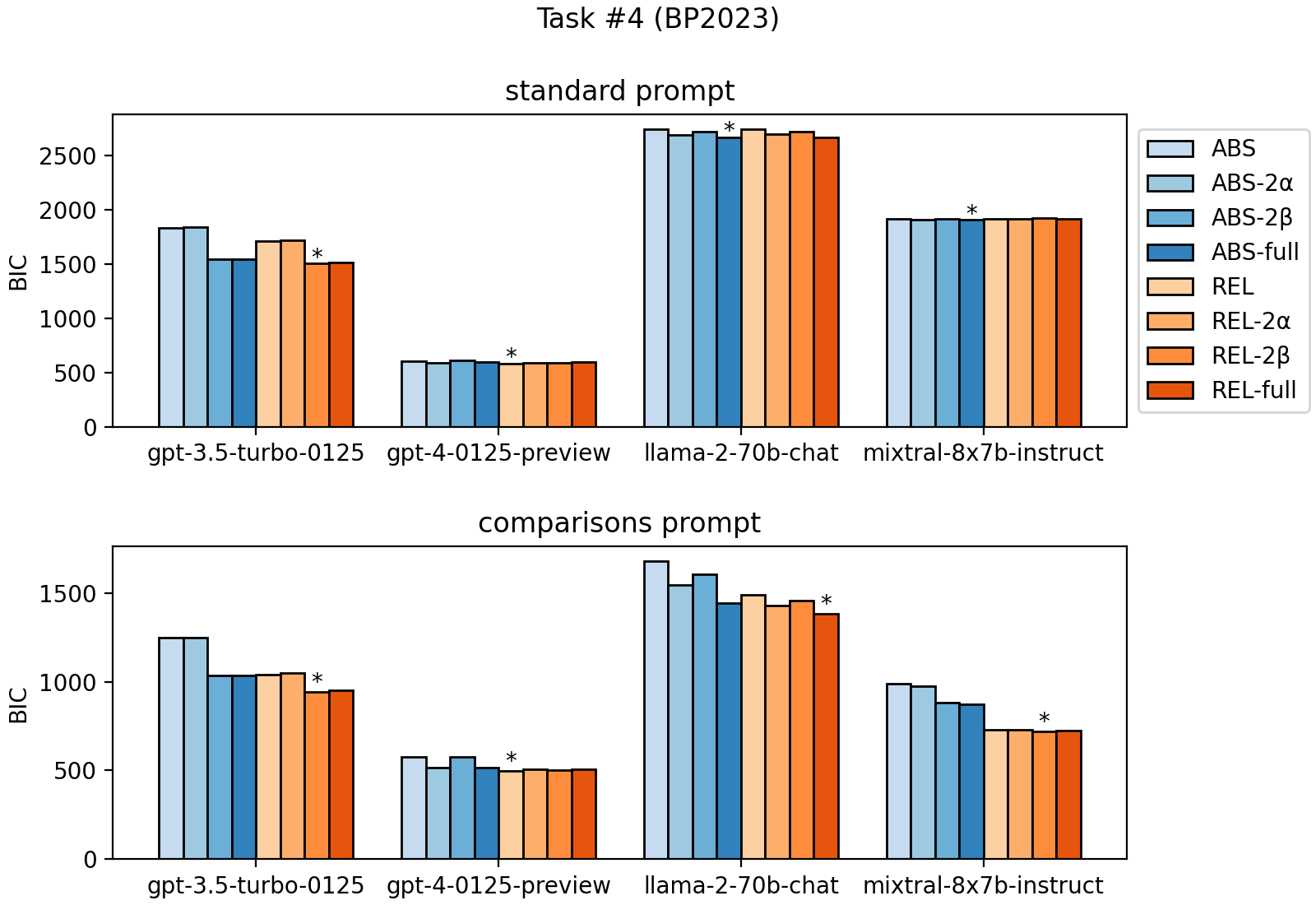}
  \caption{Model BICs in the BP2023 task. Asterisks (*) designate the best-fitting models.}
  \label{Fig10}
\end{figure}

\begin{figure}[h]
  \centering
  \includegraphics[width=5in]{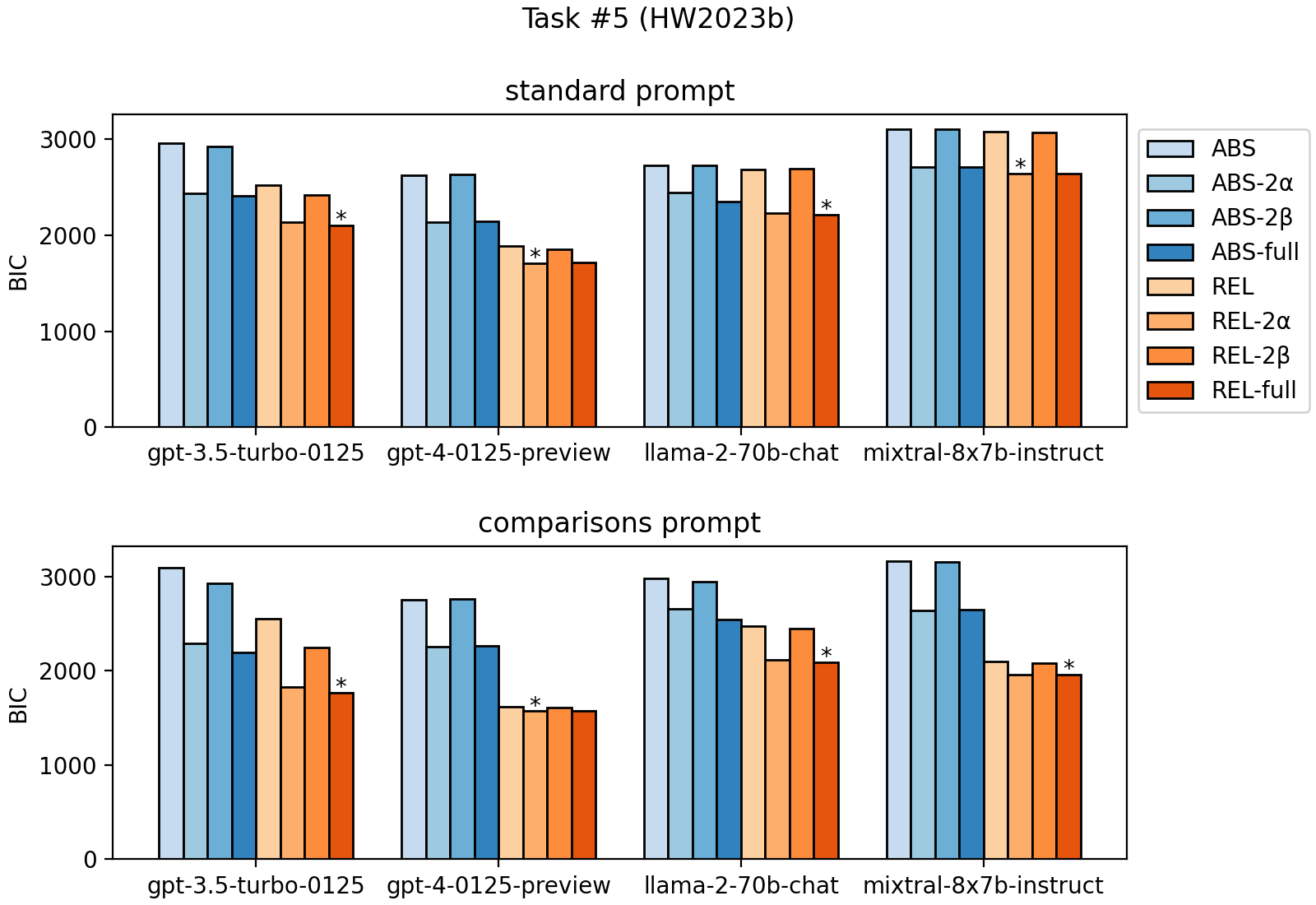}
  \caption{Model BICs in the HW2023b task. Asterisks (*) designate the best-fitting models.}
  \label{Fig11}
\end{figure}

\clearpage

\section{Task-specific choice patterns and model fit}
\label{AppendixF}

The plots below show the learning curves and transfer choice rates in each task, along with the fit of the REL-full model. Learning curves were plotted by computing the proportion of correct (i.e., reward-maximizing) choices across runs on each trial for each training context. Transfer choice rates were computed as the number of times an option was chosen divided by the number of times it was available to choose.  

\begin{figure}[h]
  \centering
  \includegraphics[width=5.5in]{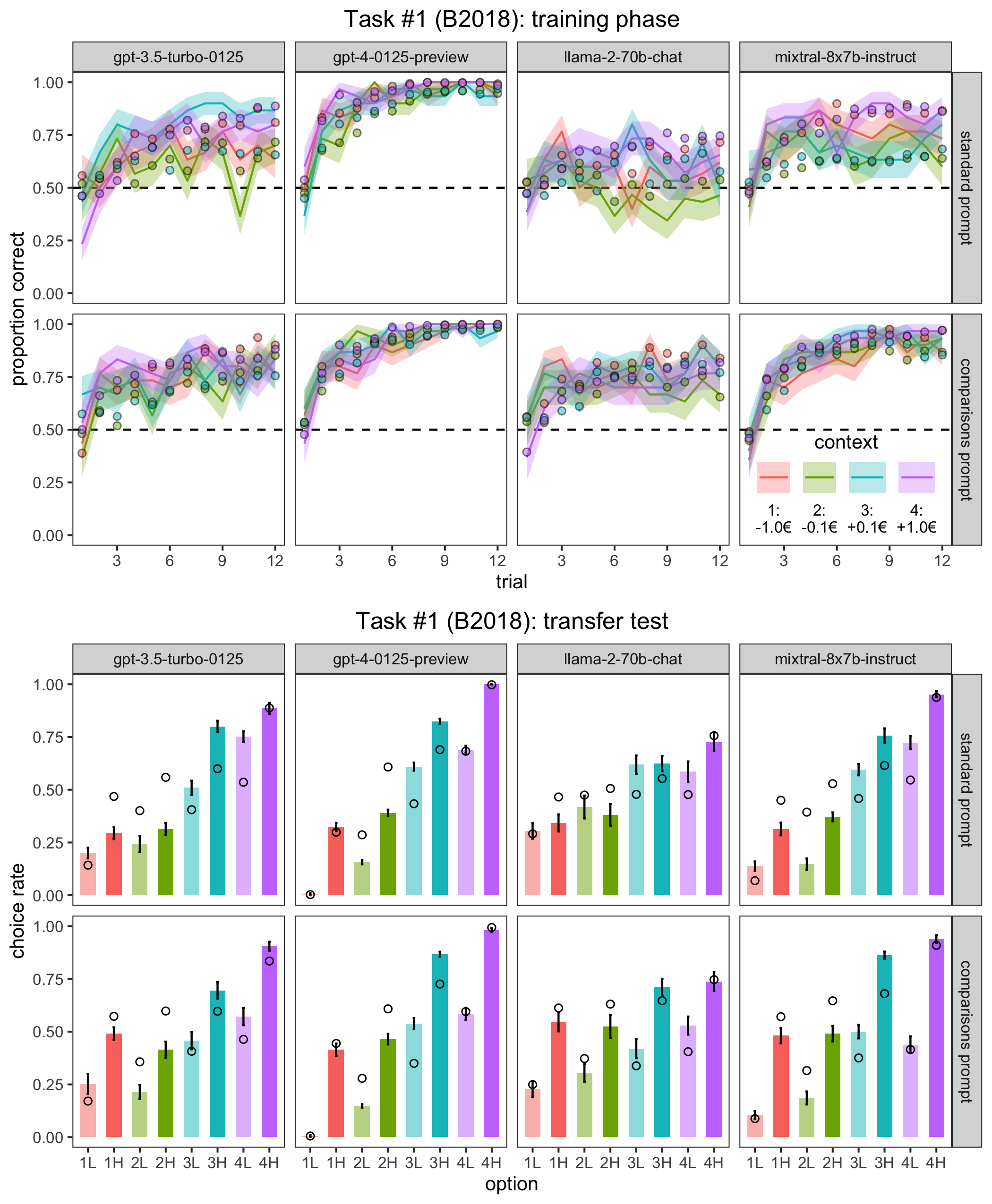}
  \caption{(a) Proportion of correct choices across training phase trials. Lines show the empirical data (+/- 1 standard error). Points show the fit of the REL-full model. (b) Mean choice rates for each option in the transfer test. Bars show empirical data (+/- 1 standard error). Points show the fit of the REL-full model.}
  \label{Fig12}
\end{figure}

\begin{figure}[h]
  \centering
  \includegraphics[width=5.5in]{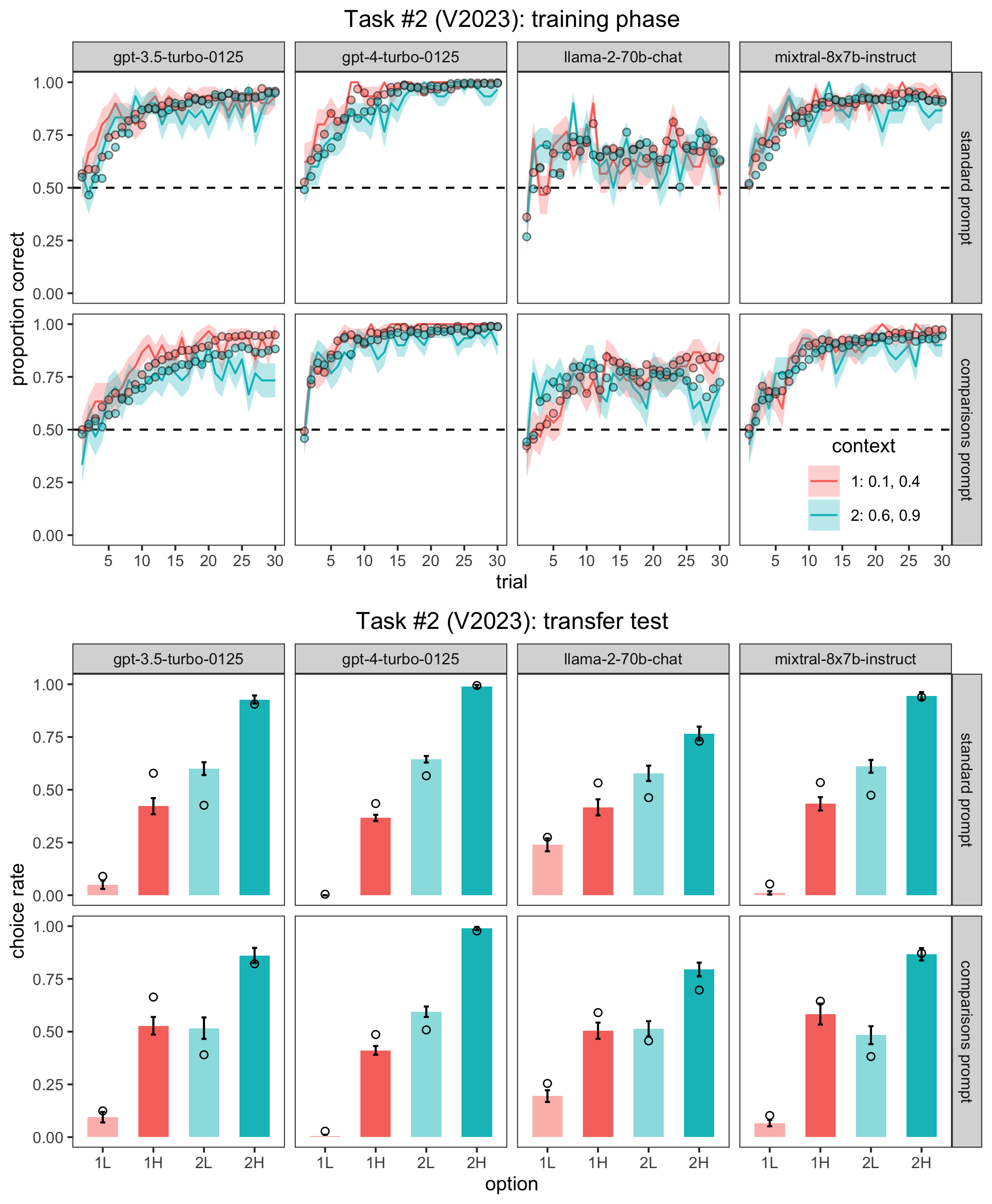}
  \caption{(a) Proportion of correct choices across training phase trials. Lines show the empirical data (+/- 1 standard error). Points show the fit of the REL-full model. (b) Mean choice rates for each option in the transfer test. Bars show empirical data (+/- 1 standard error). Points show the fit of the REL-full model.}
  \label{Fig13}
\end{figure}

\begin{figure}[h]
  \centering
  \includegraphics[width=5.5in]{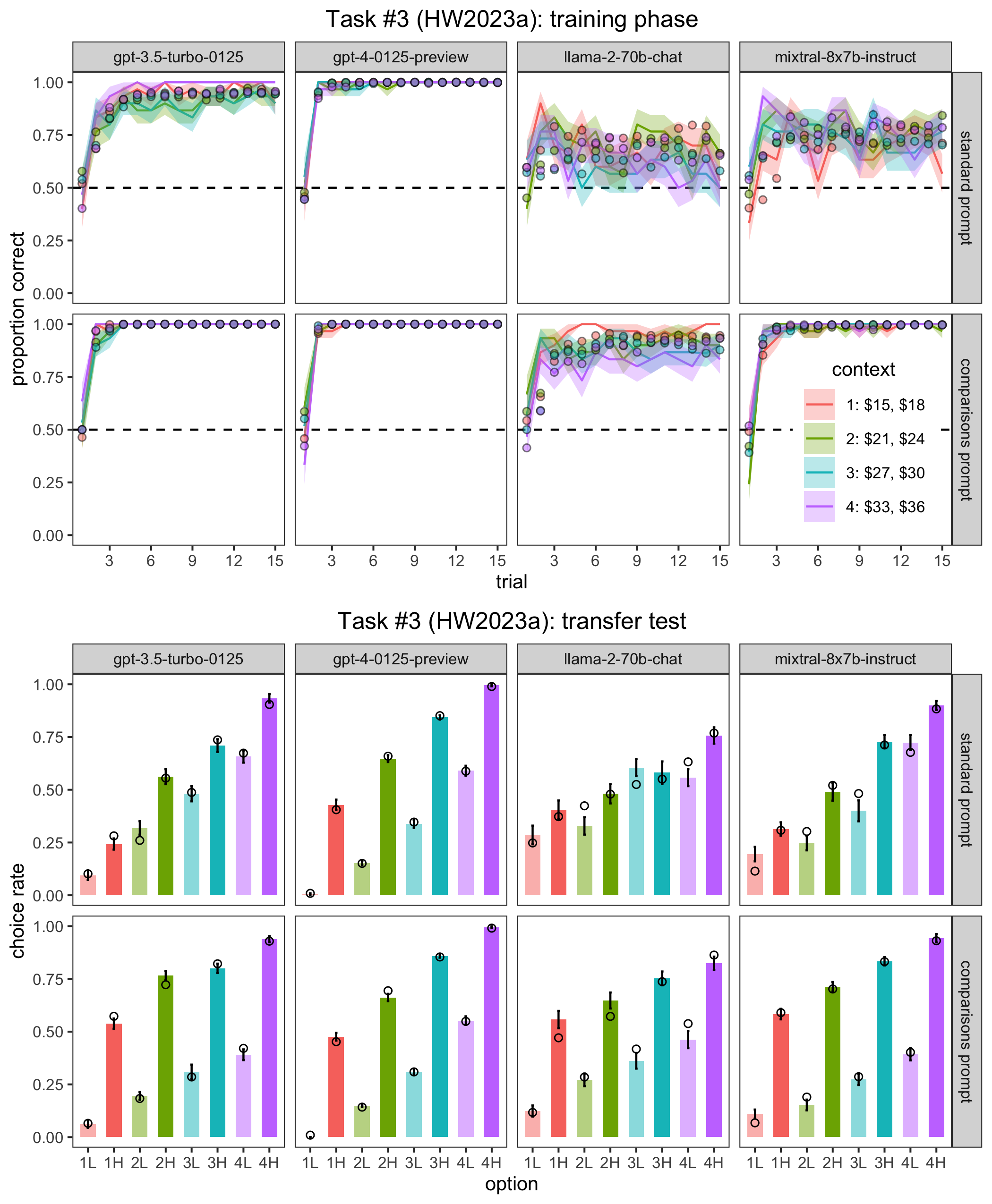}
  \caption{(a) Proportion of correct choices across training phase trials. Lines show the empirical data (+/- 1 standard error). Points show the fit of the REL-full model. (b) Mean choice rates for each option in the transfer test. Bars show empirical data (+/- 1 standard error). Points show the fit of the REL-full model.}
  \label{Fig14}
\end{figure}

\begin{figure}[h]
  \centering
  \includegraphics[width=5.5in]{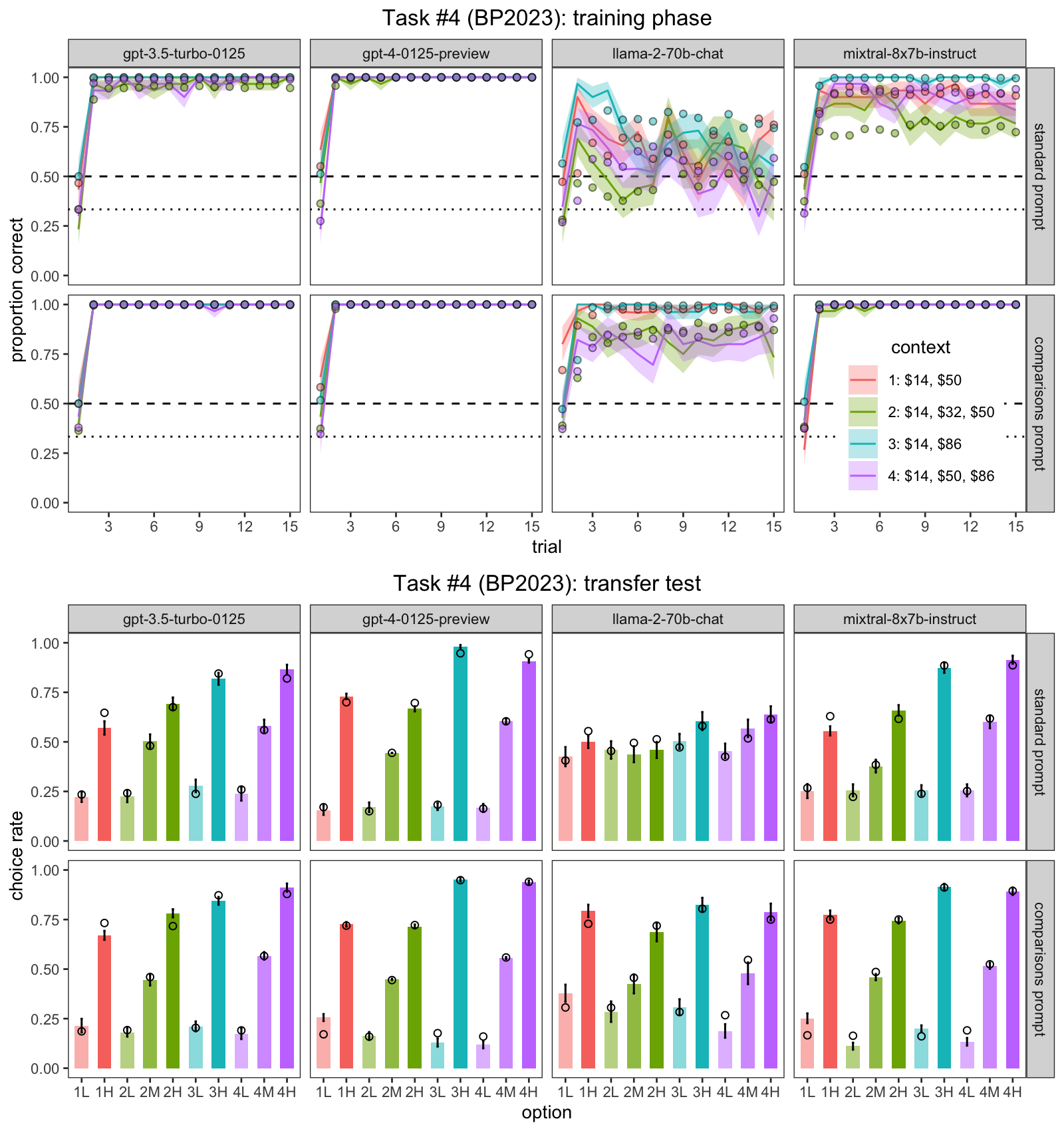}
  \caption{(a) Proportion of correct choices across training phase trials. Lines show the empirical data (+/- 1 standard error). Points show the fit of the REL-full model. (b) Mean choice rates for each option in the transfer test. Bars show empirical data (+/- 1 standard error). Points show the fit of the REL-full model.}
  \label{Fig15}
\end{figure}

\begin{figure}[h]
  \centering
  \includegraphics[width=5.5in]{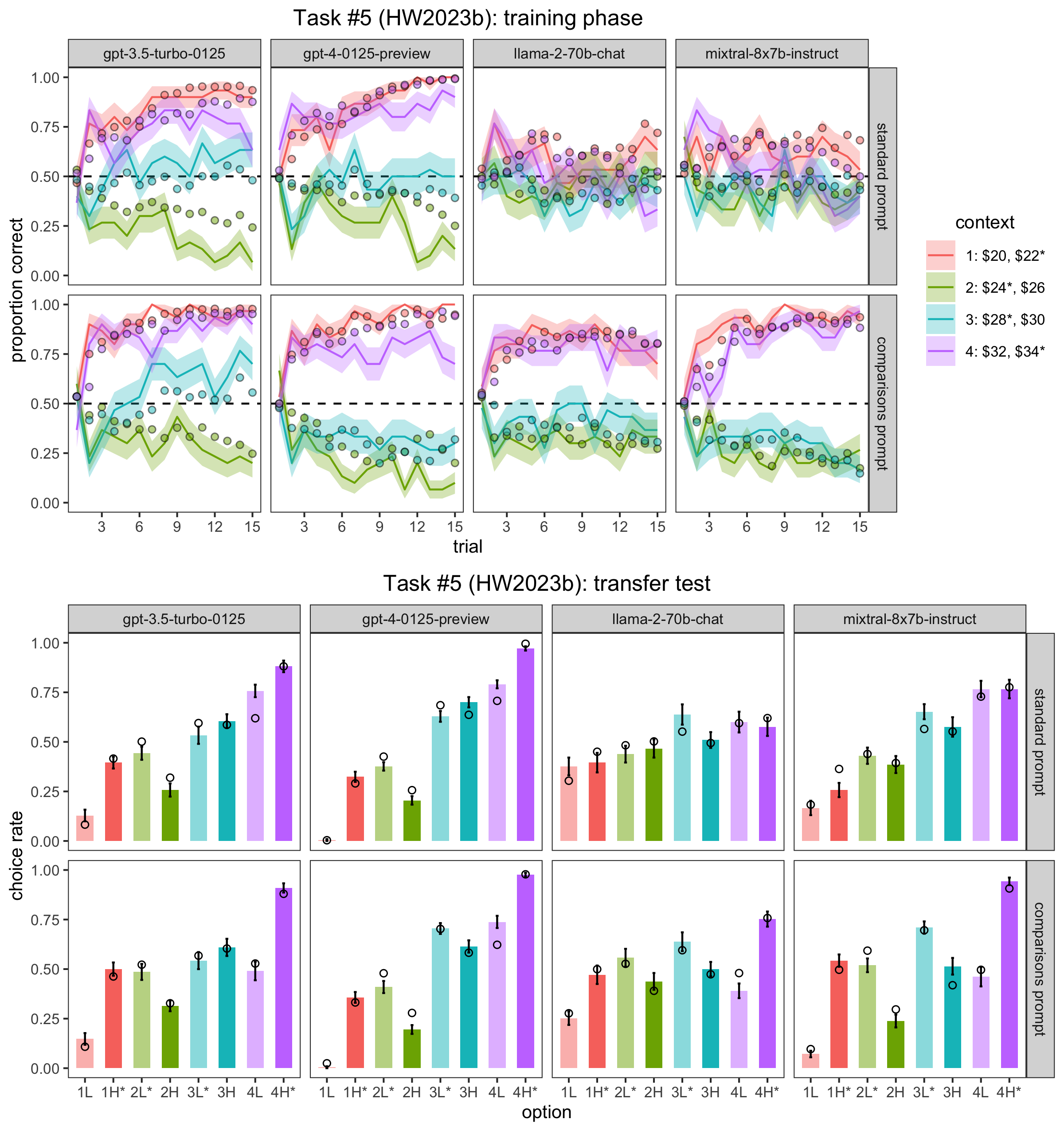}
  \caption{(a) Proportion of correct choices across training phase trials. Lines show the empirical data (+/- 1 standard error). Points show the fit of the REL-full model. (b) Mean choice rates for each option in the transfer test. Bars show empirical data (+/- 1 standard error). Points show the fit of the REL-full model. In both plots, asterisks (*) designate the options that frequently gave better relative outcomes in their original training contexts.}
  \label{Fig16}
\end{figure}

\clearpage

\section{Parameter estimates}
\label{AppendixG}

Below are the estimated parameters for the REL-full model across LLMs.

\begin{figure}[h]
  \centering
  \includegraphics[width=5.5in]{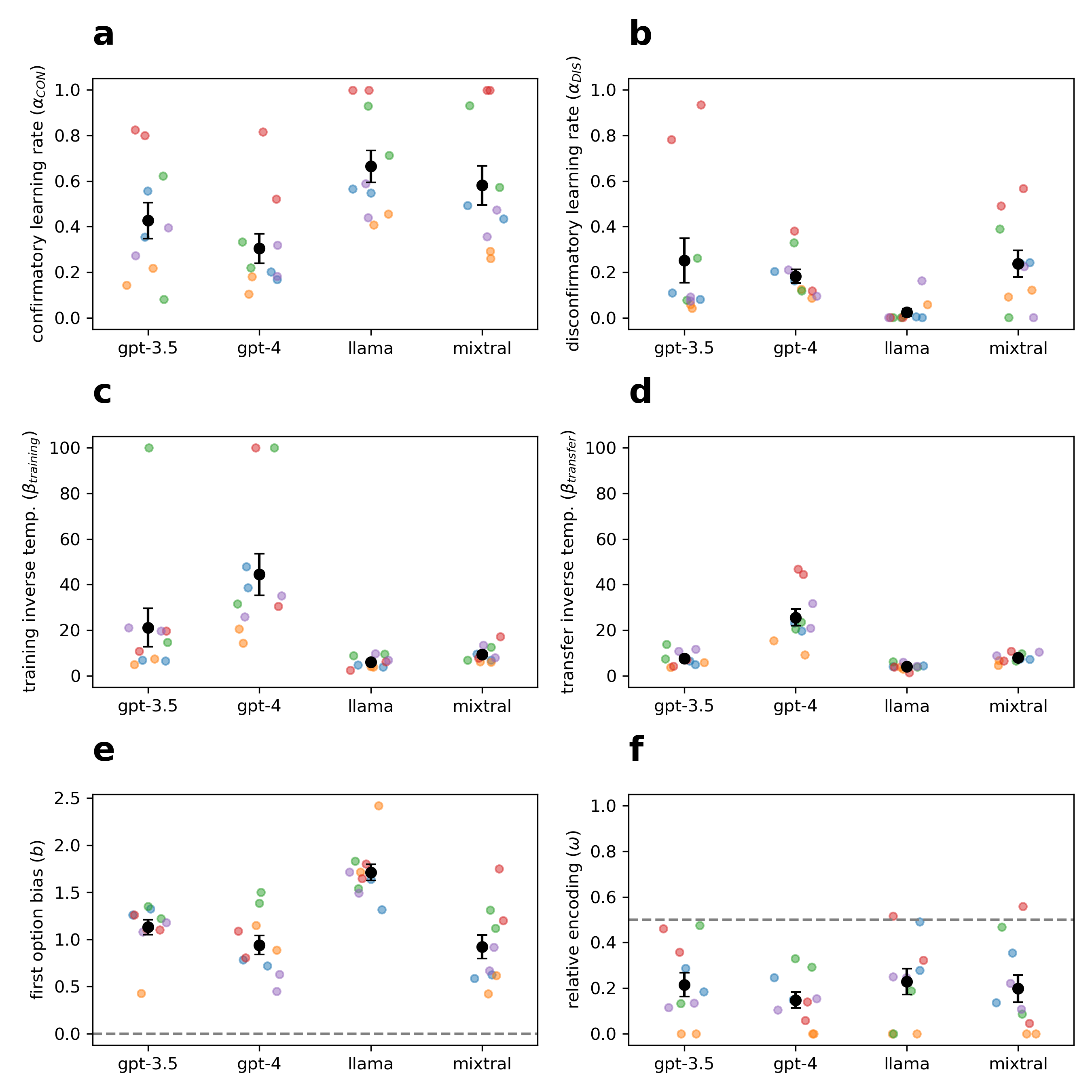}
  \caption{Parameter estimates from the REL-full model. (a-b) Learning rates. (c-d) Inverse temperatures. (e) First option bias. (f) Relative encoding. In each panel, the colored points show the estimates for a specific combination of task, prompt design, and LLM. The black points show the means and standard errors for each LLM.}
  \label{Fig17}
\end{figure}

\clearpage

\section{Additional analyses with gemma-7b}
\label{AppendixH}

The following analyses were conducted using gemma-7b, a 7 billion parameter model developed by Google \citep{team2024gemma}. We used the raw, pretrained version of the model available on \href{https://huggingface.co/google/gemma-7b}{Hugging Face}. Gemma-7b was trained on 6T tokens of mostly English text data from web documents, mathematics, and computer code. For comparison, llama-2-70b-chat was trained on 2T tokens \citep{touvron2023llama}.

\subsection{Relative value bias}

Would gemma-7b, a pretrained model with no fine-tuning, exhibit a relative value bias? To test this, we ran gemma-7b through Task \#3 (HW2023a) using a very similar procedure to what was used for the other models. Because gemma-7b is not fine-tuned for chat, the prompts were structured so that the model would essentially fill in the blank with its choice \citep{binz2023using,coda2023inducing,schubert2024context}. That is, the end of each prompt was as follows: 
\begin{tcolorbox}[colback=blue!5!white,colframe=blue!75!black,title=Prompt for gemma-7b]
\ldots\\
Q: Which slot machine do you choose?\\
A: I would choose slot machine [insert]
\end{tcolorbox}

\begin{figure}[h]
  \centering
  \includegraphics[width=5.5in]{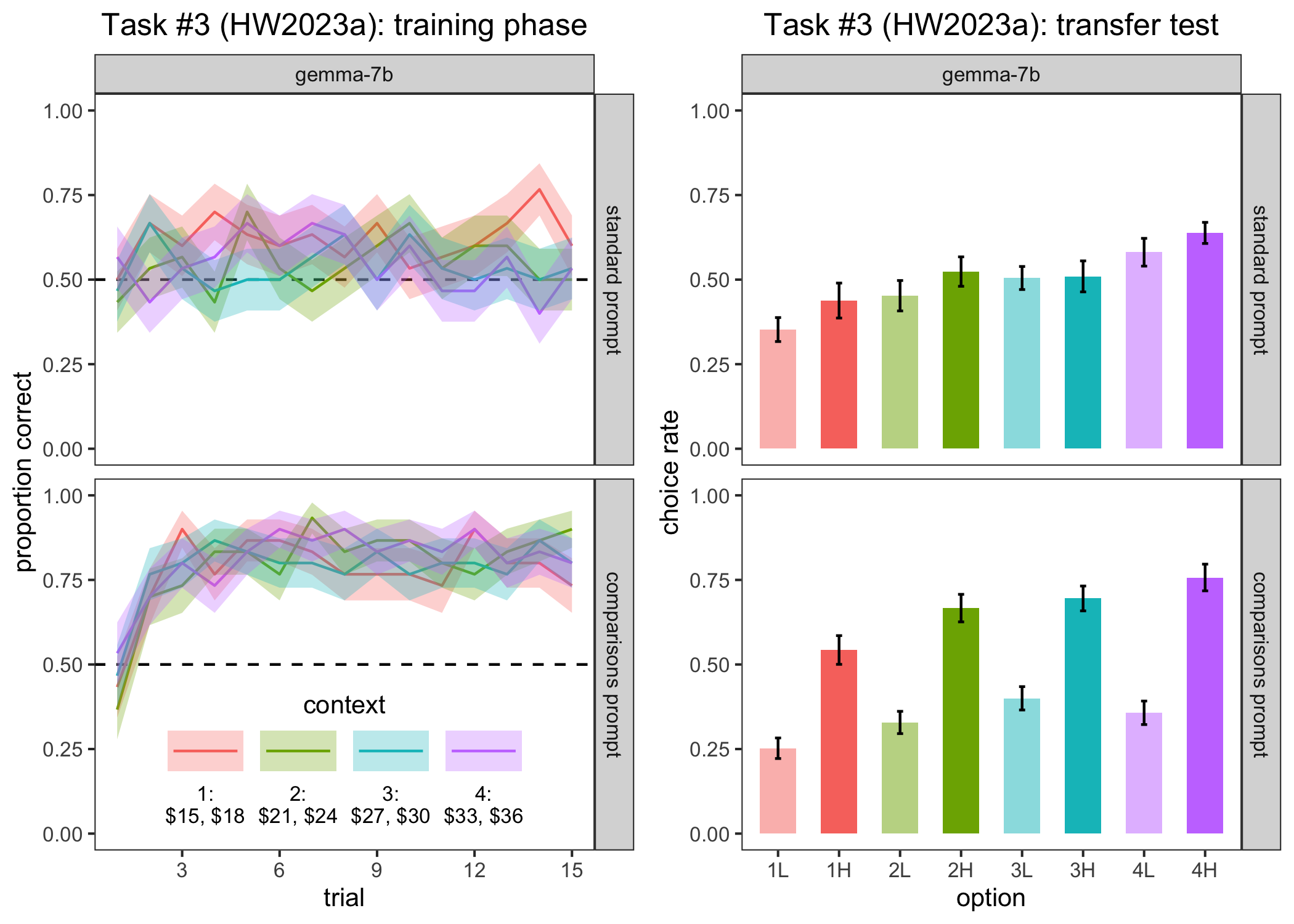}
  \caption{(left) Proportion of correct choices across training trials in both prompt conditions (+/- 1 standard error). (right) Mean choice rates for each option in the transfer test (+/- 1 standard error).}
  \label{Fig18}
\end{figure}

As shown in Figure \ref{Fig18}, gemma-7b showed weak evidence of learning with no obvious signs of relative value bias using the standard prompt. When the comparisons prompt was used, training performance was considerably improved, but relative value biases were also clearly magnified in the transfer test. This pattern closely resembles the other LLMs, especially llama-2-70b-chat (cf. Figure \ref{Fig14}), which shares many architectural features with gemma-7b.

\subsection{Analysis of hidden states}

Next, we asked whether the model's hidden layer activations might encode task-relevant information, such as the difference in absolute or relative values between choice options on each trial. We had gemma-7b perform the transfer test in the HW2023a task, on each trial recording the final hidden layer activations for the last token in the prompt (``machine''). We repeated this procedure 100 times, resulting in a 2800 $\times$ 3072 matrix of activation values (2800 trials, 3072 hidden units). The experiments were run on a single cloud A100-80G GPU.

Then, for each choice trial, we computed the difference in the absolute values of the two choice options (first minus second), as well as the difference in the relative values, both normalized between 0 and 1. For example, if the choice were between options 1H and 3L (see Figure \ref{Fig5}), the difference in absolute values would be $(18 - 27) / (36 - 15) = -0.4286$ and the difference in relative values would be $1.0 - 0.0 = 1.0$. 

Linear regression was used to predict the trial-to-trial activations in each hidden unit from the trial-to-trial differences in absolute and relative values, plus an intercept. Each hidden unit was modeled separately, for a total of 3072 regressions. An extremely conservative significance threshold was used to account for the large number of tests: The significance of each of the 6144 slope coefficients (3072 regressions $\times$ 2 predictors) was tested using a critical \textit{p}-value of $.001 / 6144 = 1.628 \times 10^{-7}$.

Using the standard prompt, 16\% of the hidden units tracked only absolute values, and 3\% tracked both absolute and relative values. In contrast, using the comparisons prompt, none of the hidden units were significantly associated with absolute values only, whereas 70\% were significantly associated with relative values only and 2\% were significantly associated with both (Table \ref{Tab6}). Clearly, the comparisons prompt had a large effect on the proportion of hidden units in the final layer that were encoding information about relative values.  

\begin{table}[h]
    \centering
    \caption{Number of hidden units with significant effects}
    \begin{tabular}{lll}
        \toprule
         & Standard prompt & Comparisons prompt \\
        \midrule
         Both non-significant & 2474 (81\%) & 873 (28\%)  \\
         Only absolute significant & 481 (16\%) & 0 (0\%) \\
         Only relative significant & 11 (0.4\%) & 2140 (70\%) \\
         Both significant & 106 (3\%) & 59 (2\%) \\
        \bottomrule
    \end{tabular}
    \label{Tab6}
\end{table}

\begin{figure}[h]
  \centering
  \includegraphics[width=5.5in]{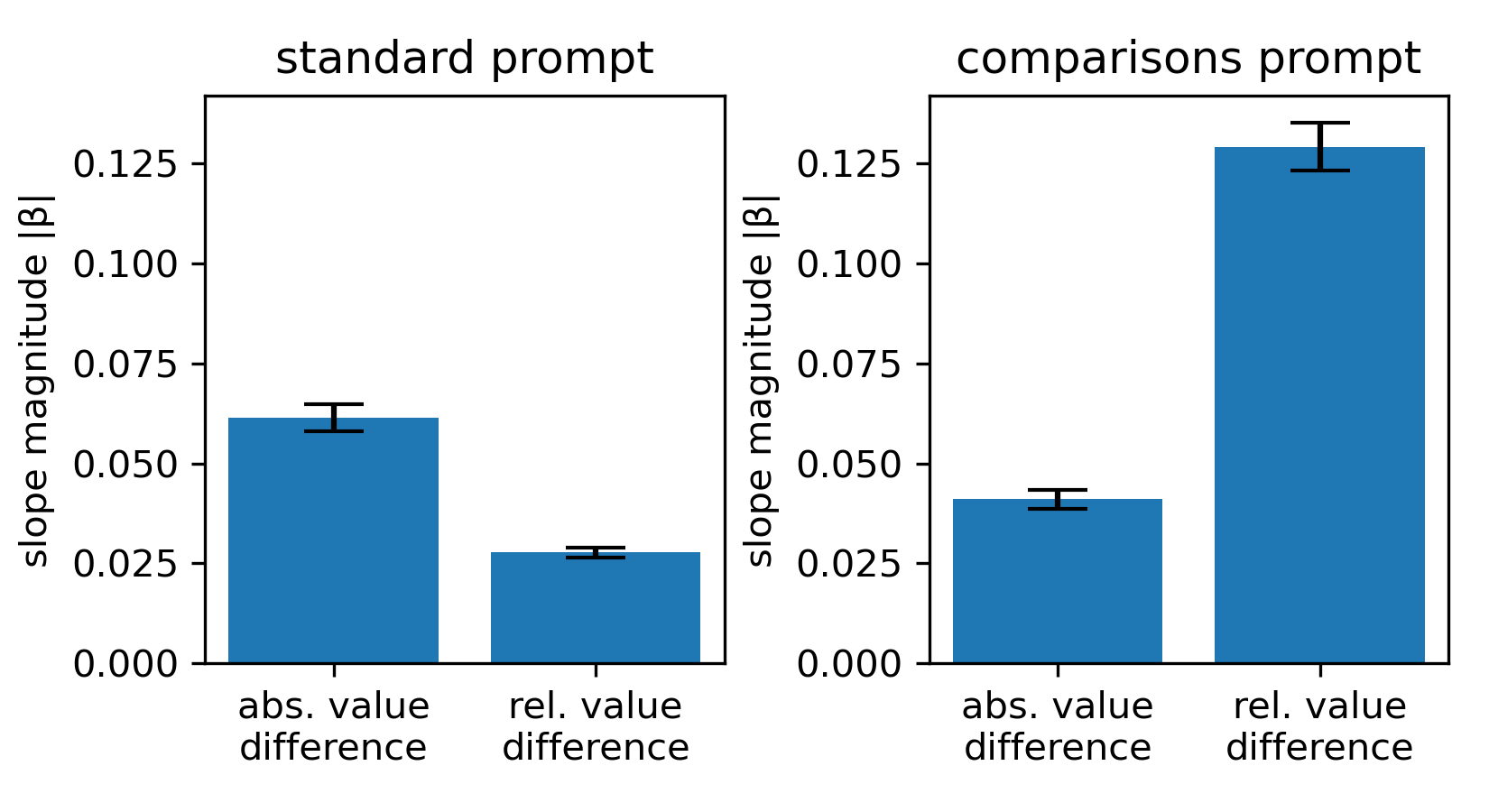}
  \caption{Trial-to-trial differences in absolute value had a stronger average effect on hidden layer activations than trial-to-trial differences in relative value using the standard prompt. The reverse was true using the comparisons prompt. Means and standard errors calculated across 3072 hidden units.}
  \label{Fig19}
\end{figure}

Since both predictors were normalized between 0 and 1, the unsigned slopes ($|\beta|$) can be interpreted as a measure of effect size. As shown in Figure \ref{Fig19}, the average effect of the absolute value difference was greater than the average effect of the relative value difference using the standard prompt, \textit{t}(3071) = 16.35, \textit{p} < .001, \textit{d} = 0.30 (paired \textit{t}-test). However, using the comparisons prompt, the relative value difference had a much larger average impact than the absolute value difference, \textit{t}(3071) = 23.83, \textit{p} < .001, \textit{d} = 0.43.

These preliminary experiments with gemma-7b demonstrate that relative value biases are observable even in the absence of fine-tuning. Further, the model's hidden layer activations appear to track task-relevant information on a trial-to-trial basis in a way that is highly dependent on the framing of the prompt. Incorporating explicit outcome comparisons in the prompt significantly enhances the representation of relative values in the model's final hidden layer. 

\end{document}